\documentclass{article}

    \PassOptionsToPackage{numbers, compress}{natbib}

\usepackage[pdftex]{graphicx} 
\usepackage{float} 
\usepackage{subfigure} 
\usepackage[export]{adjustbox}
\usepackage[ruled,vlined]{algorithm2e}

\usepackage[preprint]{neurips_2022}

\usepackage[utf8]{inputenc} 
\usepackage[T1]{fontenc}    
\usepackage{hyperref}       
\usepackage{url}            
\usepackage{booktabs}       
\usepackage{amsfonts}       
\usepackage{nicefrac}       
\usepackage{microtype}      
\usepackage{soul}
\usepackage{amsmath}

\newcommand{\ourname}{Value-Consistent~}

\newcommand{\shortname}{VCR~}
\newcommand{\shortnameno}{VCR}

\newcommand{\etal}{\textit{et al}.~}
\newcommand{\ie}{\textit{i}.\textit{e}.}

\newcommand{\eg}{\textit{e}.\textit{g}.}

\title{Value-Consistent Representation Learning for Data-Efficient Reinforcement Learning}

\author{%
  Yang Yue$^{1,2}$\thanks{This work was done when Yang Yue was an intern at Sea AI Lab.} ~~
  Bingyi Kang$^{1}$ \thanks{Corresponding Author.} ~
  Zhongwen Xu$^{1}$ ~
  Gao Huang$^{2}$ ~
  Shuicheng Yan$^{1}$ ~ \\ 
  $^{1}$Sea AI Lab~~~~$^{2}$Department of Automation, BNRist, Tsinghua University \\
  \texttt{ \{yueyang, kangby, xuzw, yansc\}@sea.com, gaohuang@tsinghua.edu.cn} 
}

\begin{document}

\maketitle

\begin{abstract}

Deep reinforcement learning (RL) algorithms suffer severe performance degradation when the interaction data is scarce, which limits their real-world application.
Recently, visual representation learning has been shown to be effective and promising for boosting sample efficiency in RL.
These methods usually rely on contrastive learning and data augmentation to train a transition model for  \emph{\textbf{state}} prediction, which is different from how the model is used in RL\textemdash performing \emph{\textbf{value}}-based planning. 
Accordingly, the learned model may not be able to align well with the environment and generate consistent value predictions, especially when the state transition is not deterministic. 
To address this issue, we propose a novel method, called \emph{value-consistent representation learning} (VCR), to learn representations that are directly related to decision-making. 
More specifically, VCR trains a model to predict the future state (also referred to as the ``imagined state'') based on the current one and a sequence of actions.
Instead of aligning this imagined state with a real state returned by the environment, VCR applies a $Q$-value head on both states and obtains two distributions of action values. 
Then a distance is computed and minimized to force the imagined state to produce a similar action value prediction as that by the real state.
We develop two implementations of the above idea for the discrete and continuous action spaces  respectively. 
We conduct experiments on Atari 100K and DeepMind Control Suite benchmarks to validate their effectiveness for improving sample efficiency.
It has been demonstrated that our methods achieve new state-of-the-art performance for search-free RL algorithms. 

\end{abstract}

\section{Introduction}

An important research direction in Deep Reinforcement Learning (RL) is to improve data efficiency, which is much demanded by the wide application of deep RL techniques in real-world scenarios.
With the state-of-the-art RL algorithms, simple tasks such as video games in Arcade Learning Environment~\cite{bellemare2013ALE} require billions of frames to achieve human-level performance~\cite{badia2020agent57}. 
In real-world applications, such as robot controllers and self-driving systems, it is impractical to obtain such a huge amount of interaction due to the costly data collection process. 
To enable deep RL to go beyond virtual games and simulators, researchers explore the data efficiency issue from various perspectives, including model-based RL~\cite{hafner2019planet,hafner2019Dreamer,simple}, auxiliary tasks~\cite{jaderberg2016auxiliary,yarats2019improving, srinivas2020curl,SPR}, data augmentation~\cite{DrQ, laskin2020rad}, etc. 
A majority of the works borrow ideas from the wider deep learning community to create additional training signals that accelerate the agent training process.
Most of these techniques are not specifically tailored for RL problems and some heavily rely on extracting high-quality visual features.

\begin{figure}[t]
    \centering
    \begin{subfigure}{.3\linewidth}
    \centering
    \includegraphics[width=\textwidth]{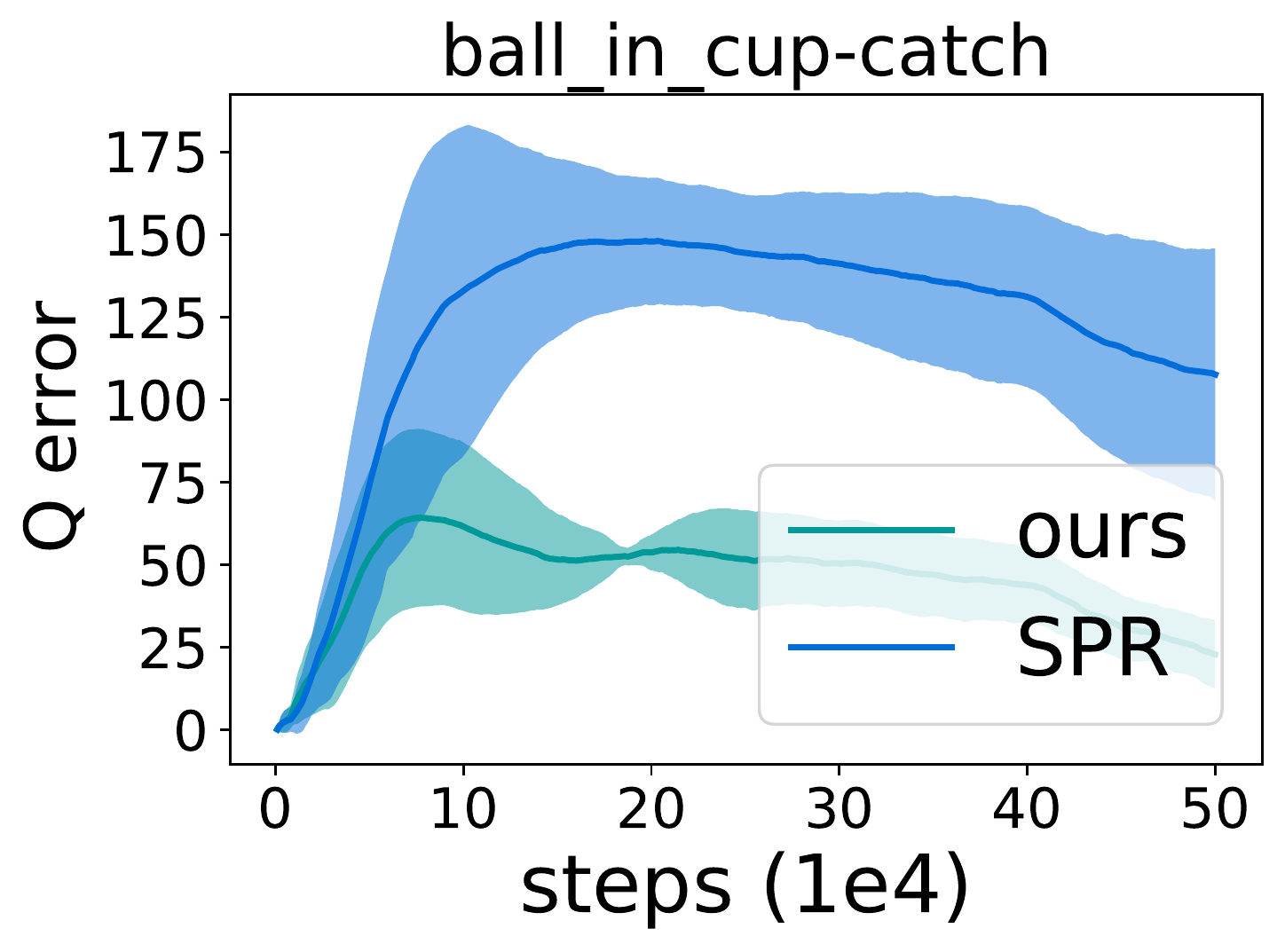}
    \end{subfigure}
    \hfill
    \begin{subfigure}{.3\linewidth}
    \centering
    \includegraphics[width=\textwidth]{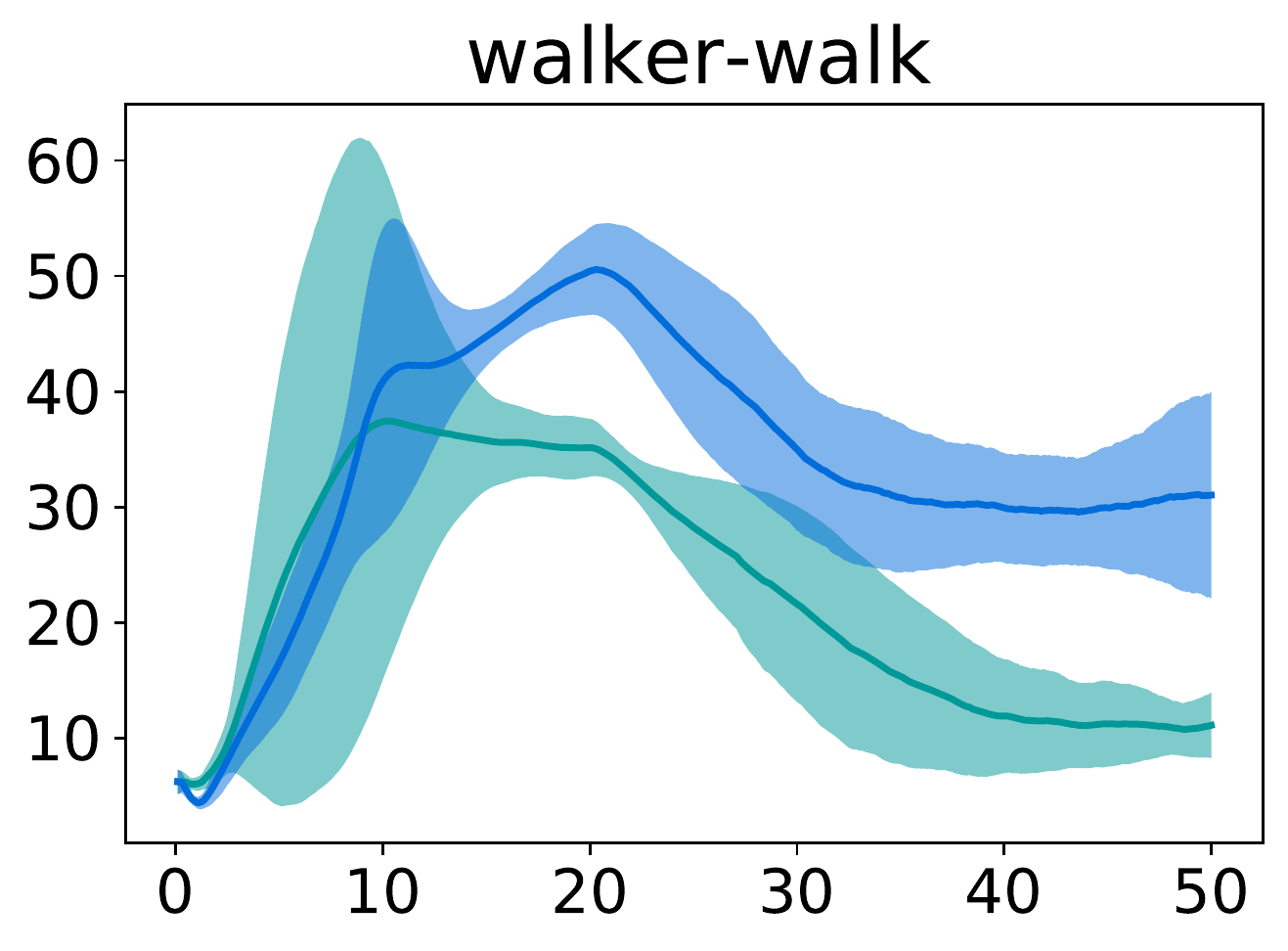}
    \end{subfigure}
    \hfill
    \begin{subfigure}{.3\linewidth}
    \centering
    \includegraphics[width=\textwidth]{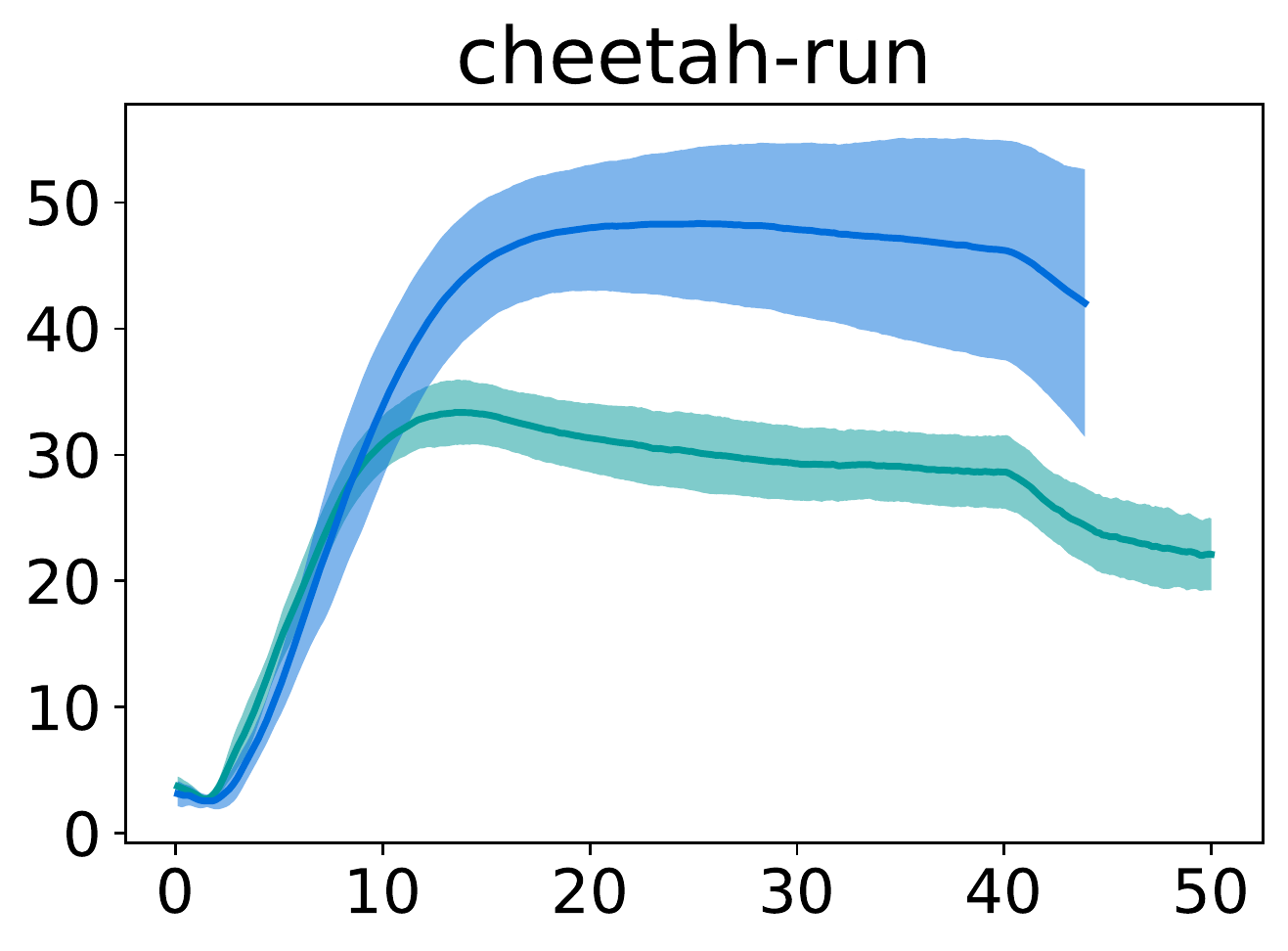}
    \end{subfigure}
    
    \begin{subfigure}{.3\linewidth}
    \centering
    \includegraphics[width=\textwidth]{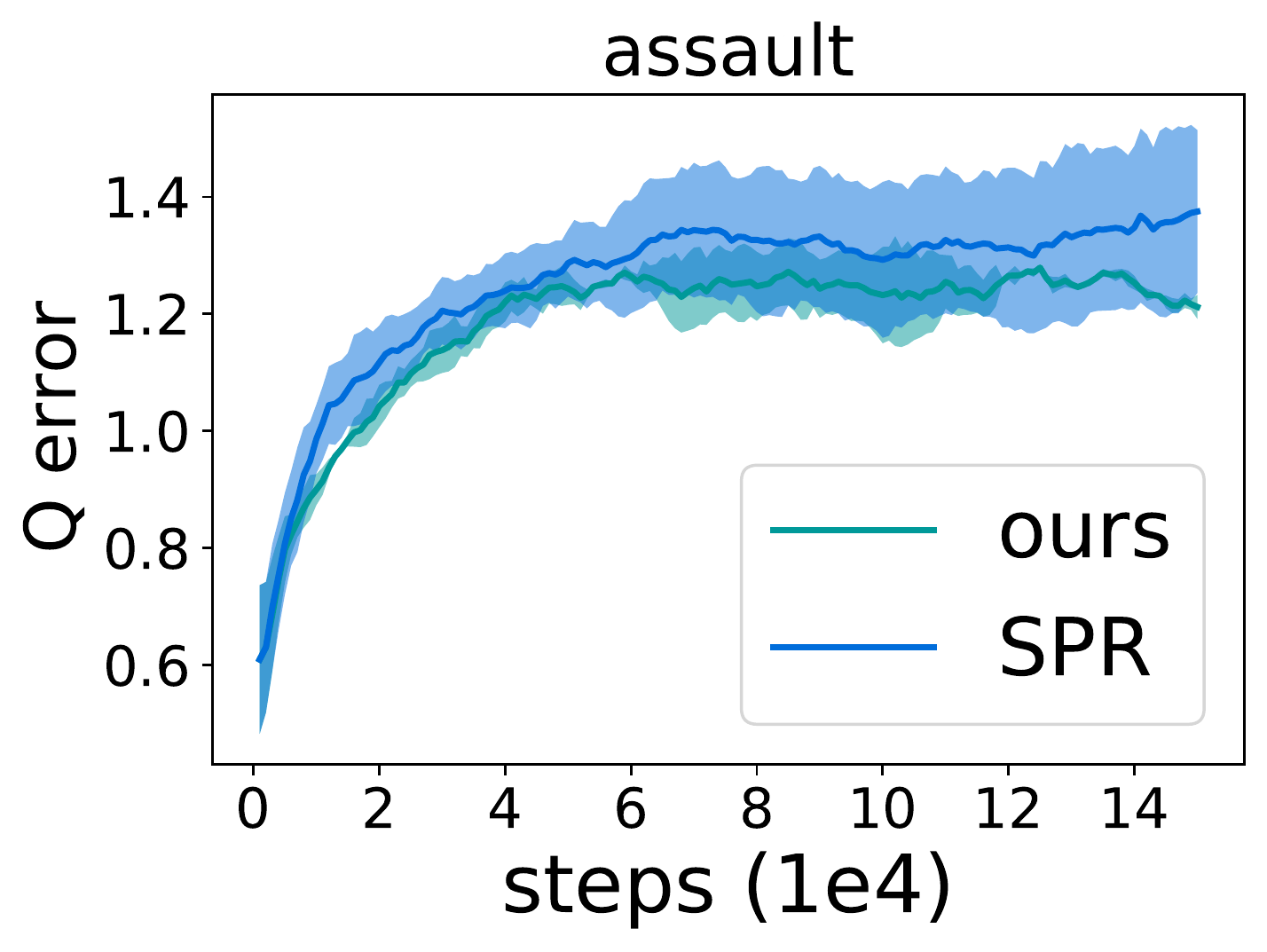}
    \end{subfigure}
    \hfill
    \begin{subfigure}{.3\linewidth}
    \centering
    \includegraphics[width=\textwidth]{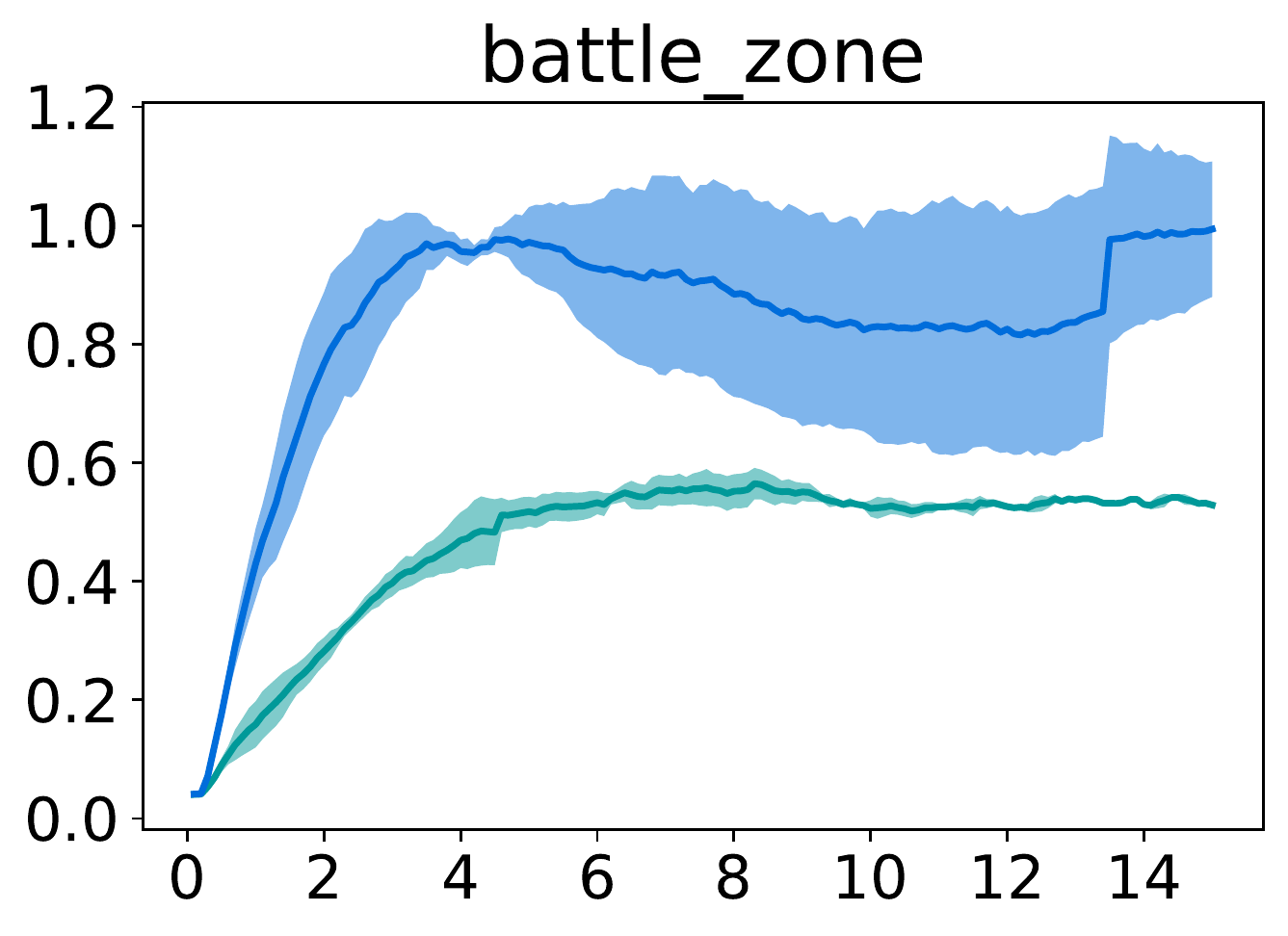}
    \end{subfigure}
     \hfill
    \begin{subfigure}{.3\linewidth}
    \centering
    \includegraphics[width=\textwidth]{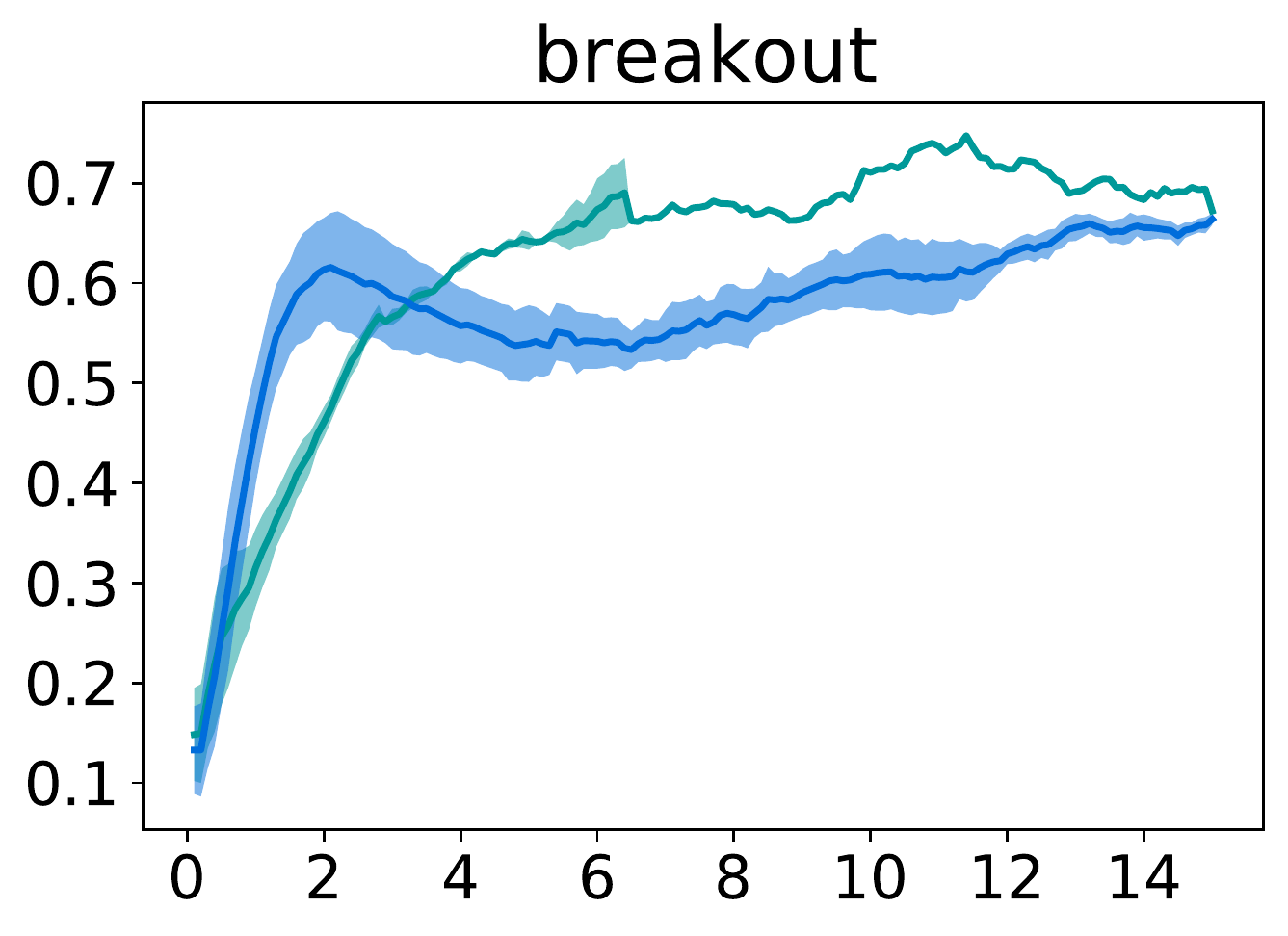}
    \end{subfigure}
    
    \begin{subfigure}{.3\linewidth}
    \centering
    \includegraphics[width=\textwidth]{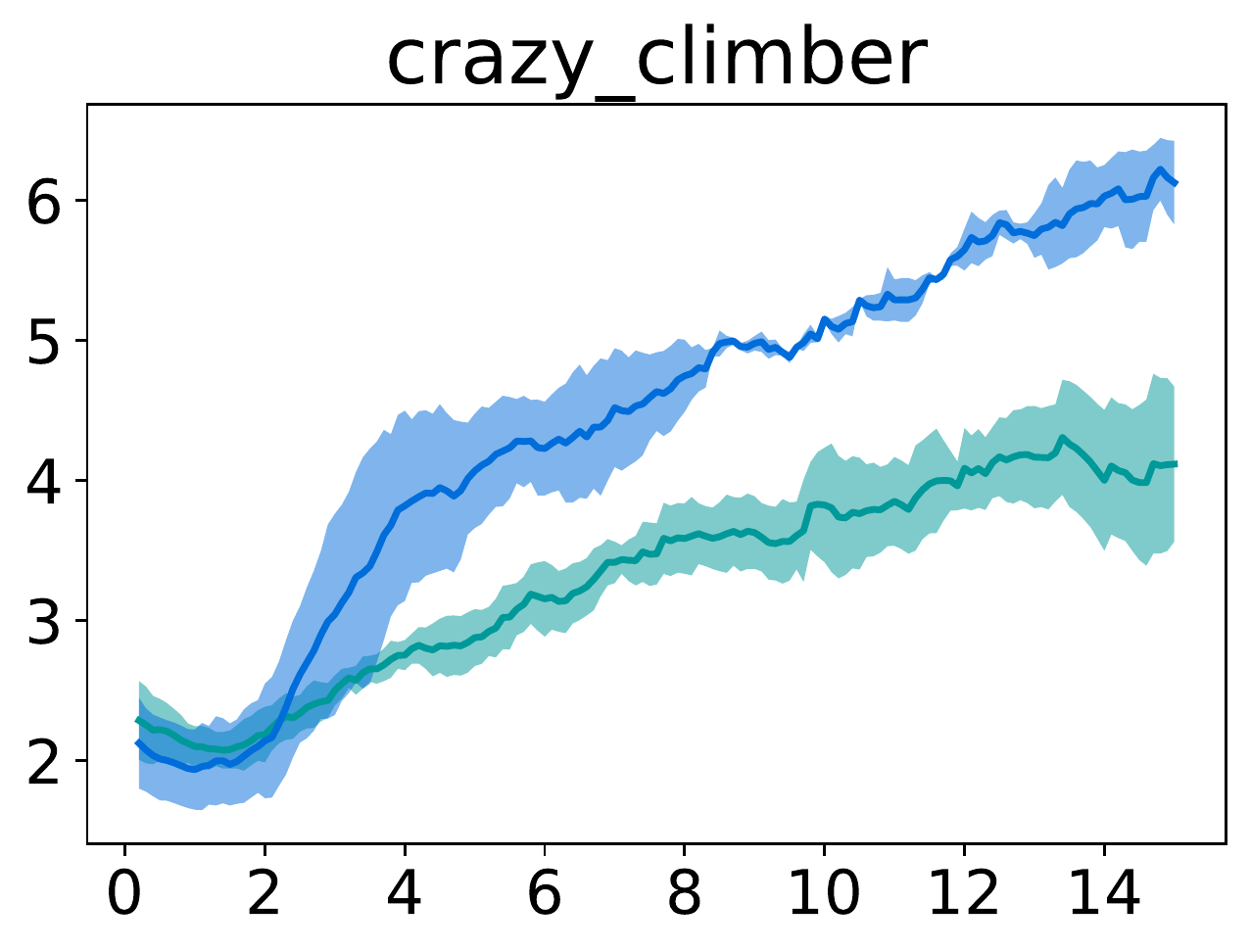}
    \end{subfigure}
    \hfill
    \begin{subfigure}{.3\linewidth}
    \centering
    \includegraphics[width=\textwidth]{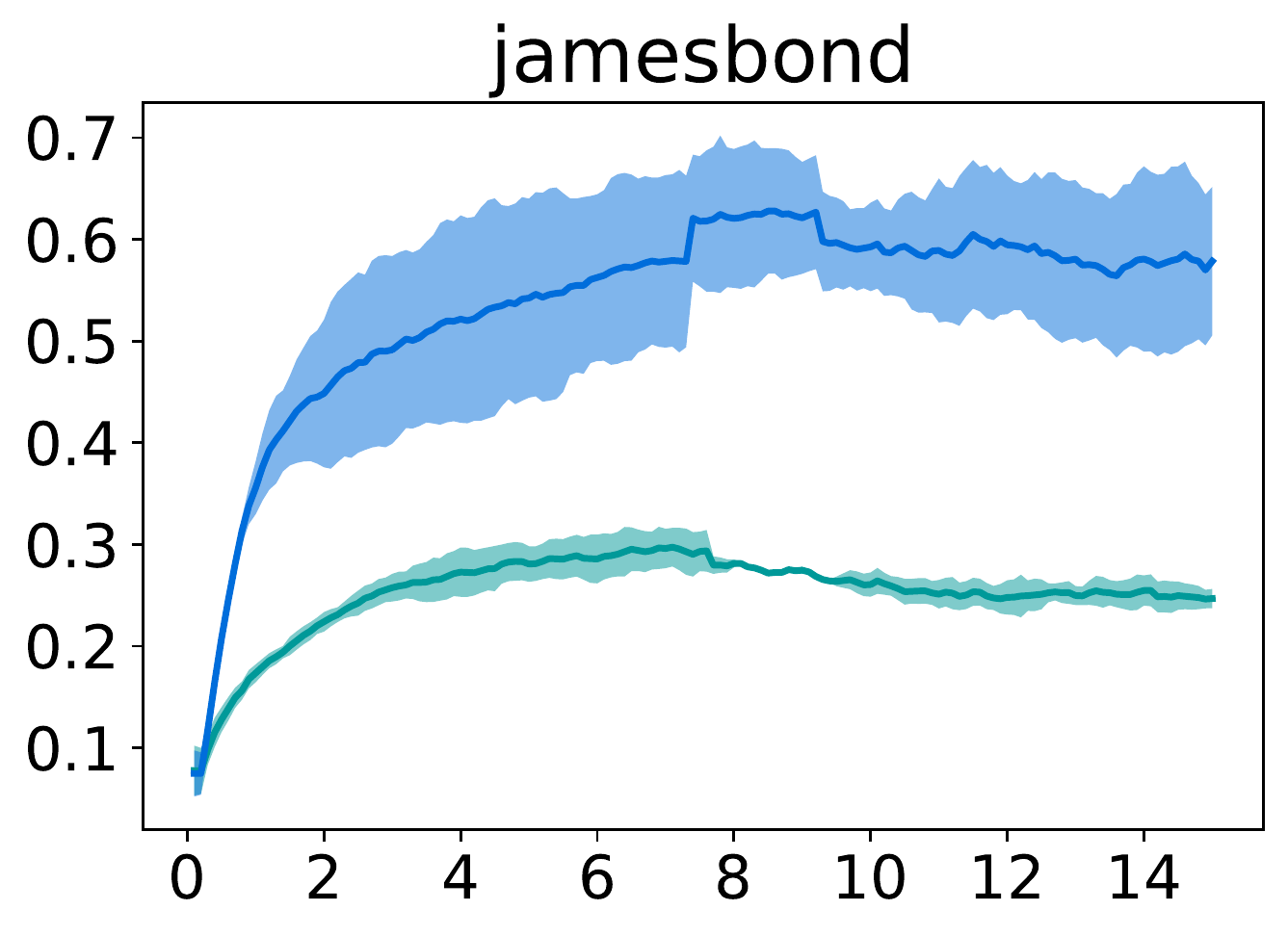}
    \end{subfigure}
    \hfill
    \begin{subfigure}{.3\linewidth}
    \centering
    \includegraphics[width=\textwidth]{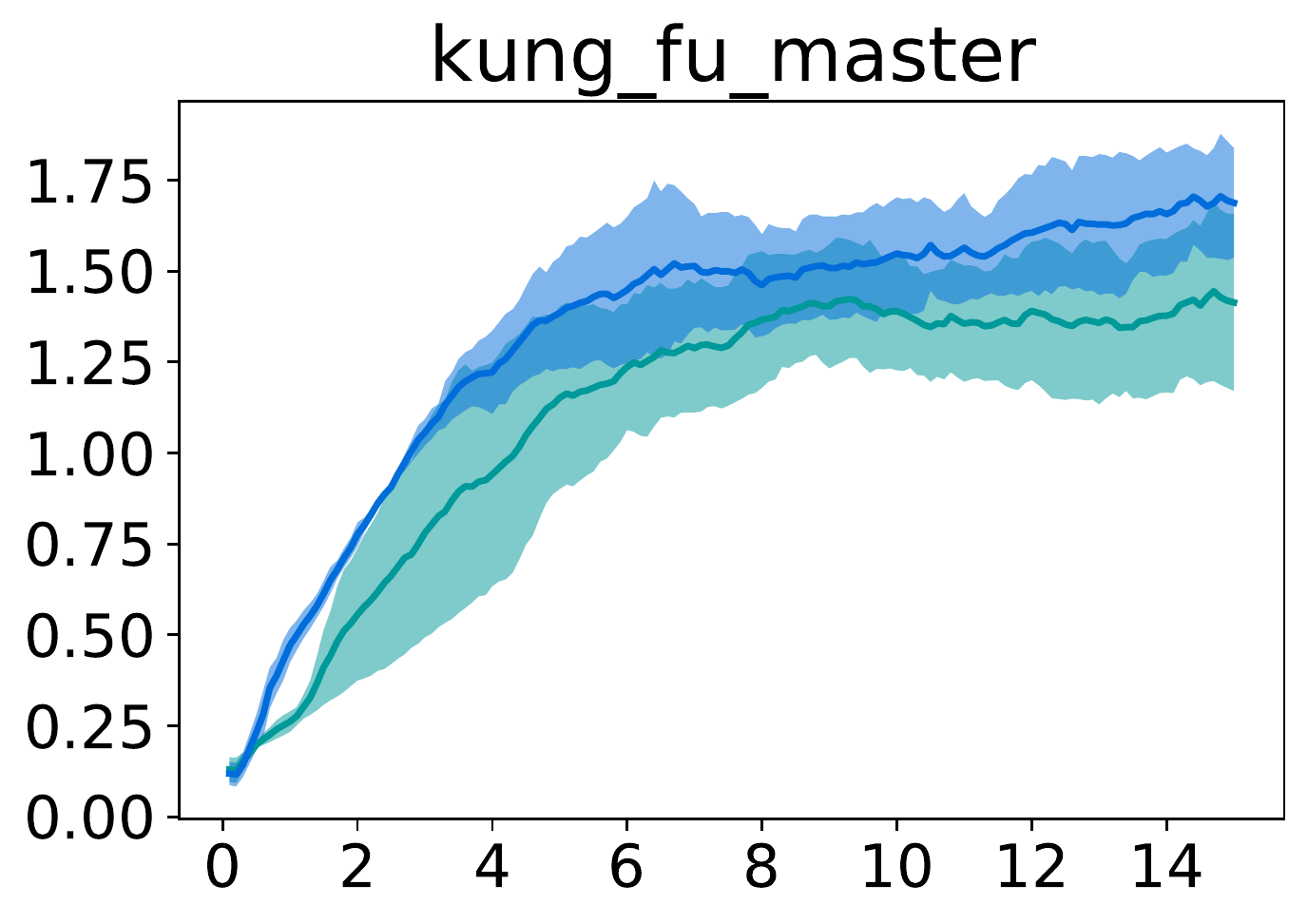}
    \end{subfigure}
    
    \begin{subfigure}{.3\linewidth}
    \centering
    \includegraphics[width=\textwidth]{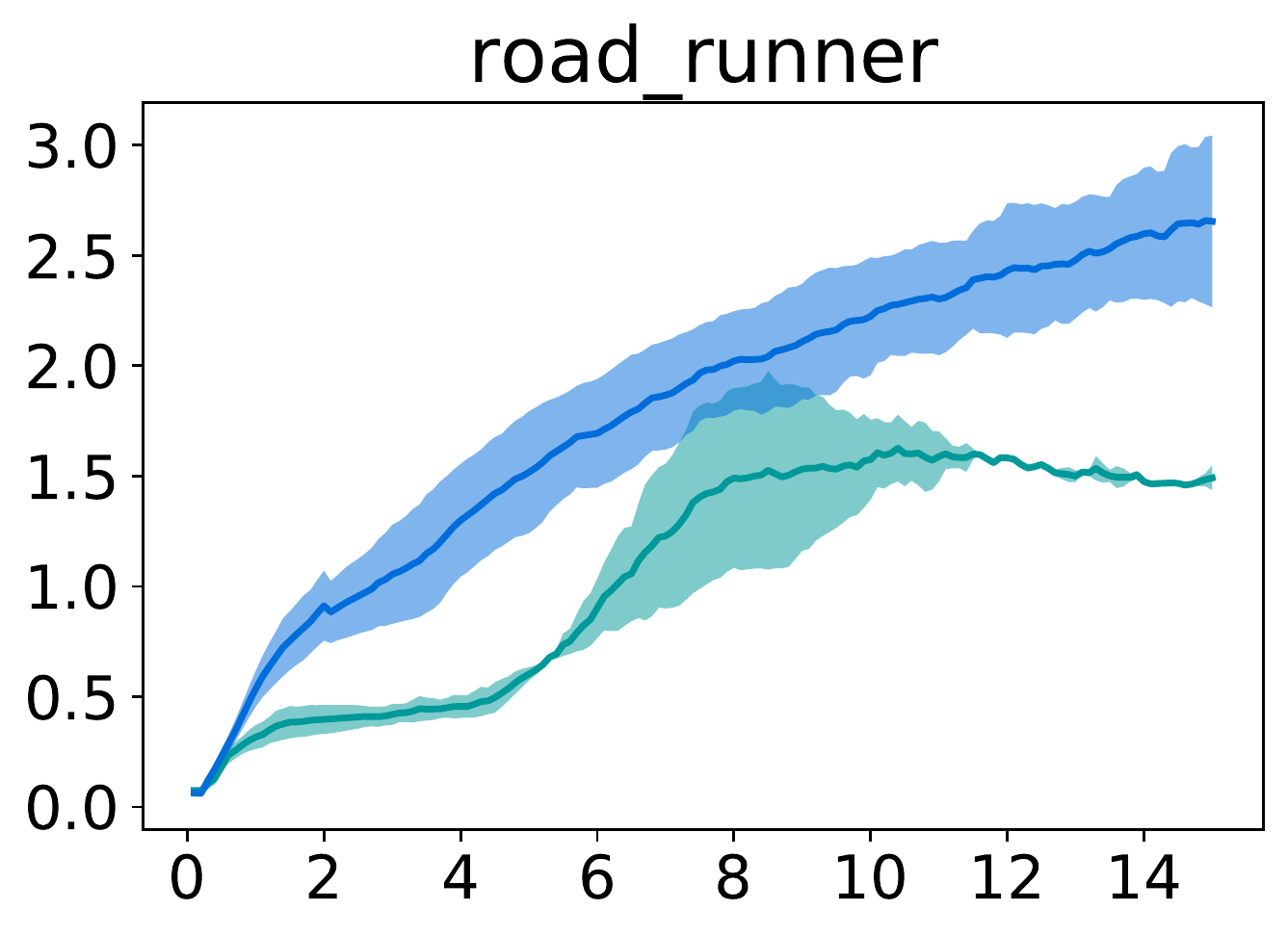}
    \end{subfigure}
    \hspace{.1\linewidth}
    \begin{subfigure}{.3\linewidth}
    \centering
    \includegraphics[width=\textwidth]{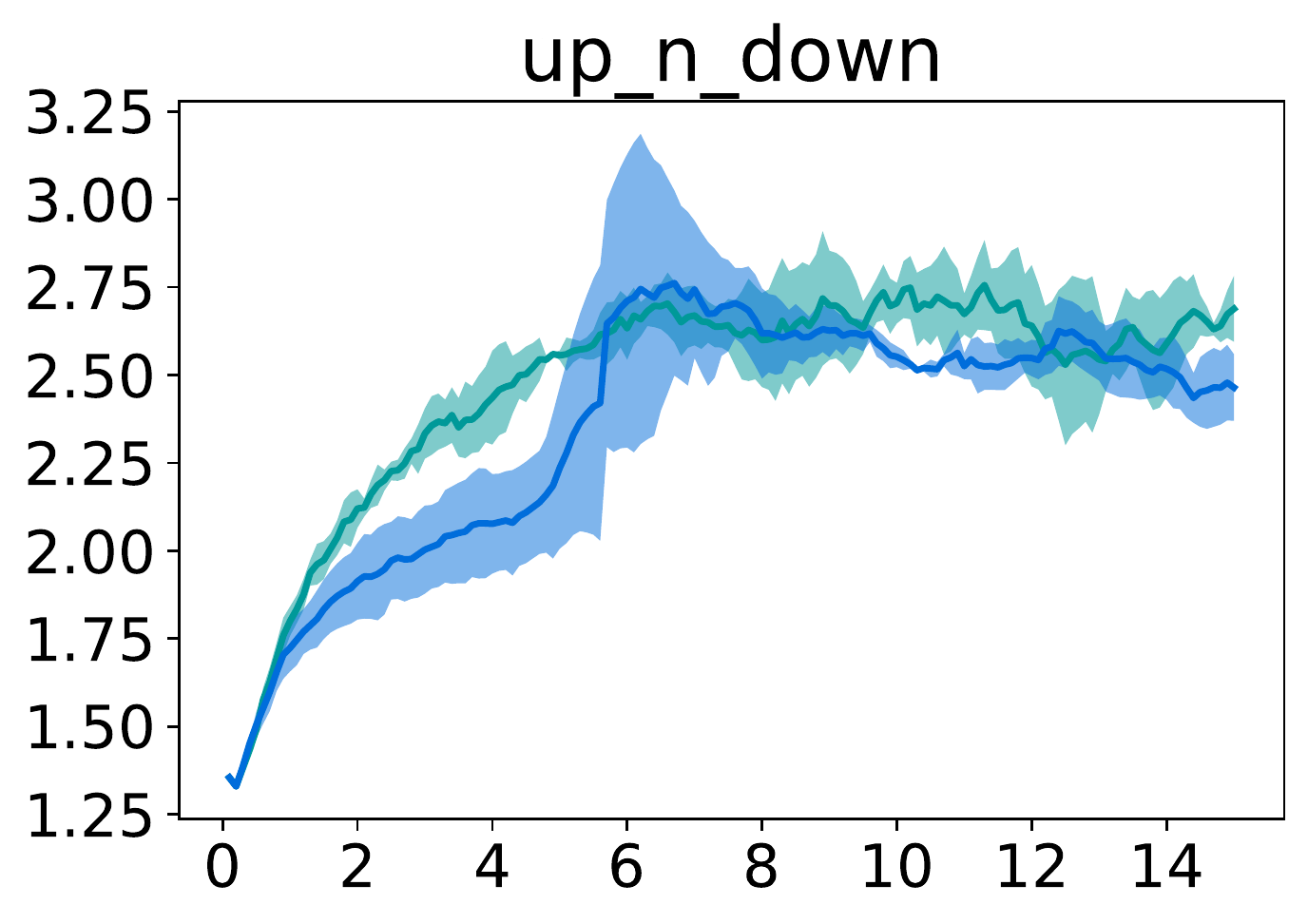}
    \end{subfigure}

    \caption{\small\textbf{$Q$ error of imagined states.} Given a dataset of real trajectories, we have a imagined state for each of the state in the dataset by shooting forward for several steps with the dynamic model. The curves represents the average error between the value estimation from imagined states and the ground-truth values along trajectories. 
    The first row shows $Q$ errors of SPR (blue) and \shortname (ours, green) on 4 DeepMind Control environments with 6 seeds. The second and third rows shows $Q$ error on a subset of Atari 100K games with 3 seeds. The shaded area represents standard deviation. Details are described in Sec.~\ref{sec:q-error}.}
    \label{fig:q-error}
\end{figure}

Recently, Self-Predictive Representation (SPR; \citeauthor{SPR} \citeyear{SPR}) introduces contrastive learning into transition model learning. Specifically, SPR aims at learning an embedding space in which the agent can predict future state embeddings. Despite its effectiveness, it solely focuses on learning discriminative state features, ignoring the fact that some information in the raw pixels is unnecessary or even distracting for decision-making. This part of information would be encoded into representation by visual contrastive learning. Moreover, it is possible in practice that two visually similar states result in significantly different returns. 
In other words, the learned representation in this visual way might be good for recognition but not optimal for estimating state value and solving the decision problem. As evidenced by Fig.~\ref{fig:q-error}, the predicted values based on imagined states of SPR are consistently deviated from the true values (by Monte-Carlo estimation) across multiple environments. 

Similar observations have been made on the Value Equivalence (VE) principle~\cite{grimm2020value} for model-based RL, which advocates that a model should be able to generate the same Bellman updates as the real environment
rather than directly modeling state-to-state transitions. The success of some state-of-the-art algorithms such as MuZero~\cite{MuZero}, Value Prediction Network~\cite{VPN} and Predictron~\cite{predictron} can be attributed to this principle. However, value equivalence is only studied with state value functions $V(S)$, and is often coupled with search-based algorithms, which makes it non-trivial to apply VE for value-based RL. 

In this paper, we develop a value-consistent metric for action-values (\ie, $Q$-values) and propose a novel method called \emph{Value-Consistent Representation Learning} (VCR) to boost sample efficiency for action-value based RL methods.
In contrast to existing data-efficient RL ideas, VCR is based on RL semantics other than losses constructed purely upon input states.
Specifically, we introduce a dynamics model to predict the next states in a latent representation space induced by a state encoder. When a real trajectory from the environment is provided, our dynamics model can roll out an imagined trajectory from the initial state by taking the same sequence of actions. Then, for each of the real and imagined state pairs, we obtain two $Q$-value distributions (over all available actions). A value-consistent loss function is applied to align these two distributions. VCR can be viewed as a clean implementation of value equivalence with action-values so that it is highly extendable in a significant family of RL algorithms. The idea is simple yet effective and it can be easily integrated into any value-based RL algorithm. We demonstrate the effectiveness of the idea with two concrete implementations based on Rainbow DQN~\cite{hessel2018rainbow} and Soft Actor-Critic (SAC)~\cite{haarnoja2018sac}.

We conduct experiments to validate the effectiveness of VCR on two benchmarks: Atari 100K~\cite{simple} (discrete action) and DeepMind Control Suite~\cite{tassa2018DMC} (continuous action). Despite its simplicity, the results show our method can boost sample efficiency significantly and achieve new state-of-the-art in search-free methods.
\section{Related Work }

\subsection{Data-Efficient Reinforcement Learning}
Reinforcement learning algorithms suffer severe performance degradation when only a limited number of interactions are available. 
There are various methods trying to tackle this problem, by either model-based~\cite{simple} or model-free learning~\cite{DER,OTR,DrQ}.
For example, 
SimPLe~\cite{simple} utilizes a world model learned with collected data to generate imagined trajectories, which are combined with real trajectories to train the agent. 
Later, Data-Efficient Rainbow~\cite{DER} and OTRainbow~\cite{OTR} show that Rainbow DQN with hyperparameter tuning can be a strong baseline for the low-data regime by simply increasing the number of steps in multi-step return and allowing more frequent parameter update.

Recently, leveraging computer vision techniques is drawing increasingly more attention from the community to boost representation learning in RL.
DrQ~\cite{DrQ} makes a successful attempt by introducing image augmentation into RL tasks where an agent takes visual inputs, while Yarats \etal~\cite{yarats2019improving} use image reconstruction as an auxiliary loss function. 
Further, rQdia~\cite{lerman2021rqdia} exploit the idea that $Q$-values should be invariant across different augmentations of the same image with arbitrary action.
Inspired by the remarkable success of contrastive learning for representation learning~\cite{SimCLR, BYOL}, the contrastive loss has been integrated into RL as an effective component.
For example, CURL~\cite{srinivas2020curl} forces different augmentations of the same state to produce similar embeddings and different states to generate dissimilar embeddings, with the contrastive loss jointly optimized with an RL loss.

Recently, SPR~\cite{SPR} and KSL~\cite{KSL} make a sophisticated design by employing contrastive loss in transition model learning. In this way, an agent can learn representations that are predictable when the previous state and action are given.
As a follow-up, PlayVirtual~\cite{yu2021playvirtual} introduces a backward prediction model that enables the agent to imagine forward and backward to form a cycle. Therefore, arbitrary actions can be taken in imagination to compute cycle-consistent loss.
EfficientZero~\cite{efficientzero} introduces SPR loss into MuZero~\cite{MuZero}, achieving super-human performance on the Atari 100K for the first time. 

These successes demonstrate that self-predictive representation learning is indeed a promising way to improve sample efficiency. We also choose to base our method on SPR. However, these methods only focus on making accurate state predictions, ignoring that value prediction is key to decision making. 
In this work, we propose value-consistent representation learning and show its importance for decision-making.

\subsection{Value Equivalence Principle}

Some works in model-based RL have proposed a high-level idea of learning the transition model in terms of the value space.
Value-Aware Model Learning (VAML)~\cite{VAML, farahmand2018iterative} incorporates the knowledge of value function into optimization to learn the probabilistic transition model in model-based reinforcement learning.
Grimm \etal~\cite{grimm2020value} come up with the value equivalence principle which forces the learned transition model to have the same Bellman operator updates with the real environment model conditioned on a set of value functions and policies. 
Several empirically successful works like VPN~\cite{VPN}, VIN~\cite{VIN} and MuZero~\cite{MuZero} can be viewed as the instances that follow the value equivalence principle when considering diverse forms of Bellman operators and value approximations.
Self-consistent models and values~\cite{farquhar2021self} boosts the performance of Muesli agent~\cite{Muesli} by encouraging a value function to satisfy the Bellman equation under a learned transition model and a learned policy.


Although based on similar motivation as the value equivalence principle, our core idea greatly differs from the above works. All above works lie in the category of planning-based RL and leverage value regularization or value equivalence to train a dynamics model for planning. However, our method, a search-free algorithm, harnesses value consistency to learn a transition model for better state representation. Besides, our method employs the reward from the real environment to estimate the target value, while those planning algorithms use predicted rewards with an extra reward predictor.
\section{Method} 
We are especially interested in Reinforcement Learning (RL) in a low-data regime, where only a limited number of interactions are allowed. 
In this section, we detail our value-consistent representation (\shortnameno) learning method step by step. First, we prepare the readers with certain preliminaries required by \shortnameno. Then, we introduce the intuition and the overall framework for \shortnameno. In the end, we provide two implementations of VCR for both discrete and continuous action settings. 

\begin{figure}[t]
    \centering
    \includegraphics[width=0.48\textwidth]{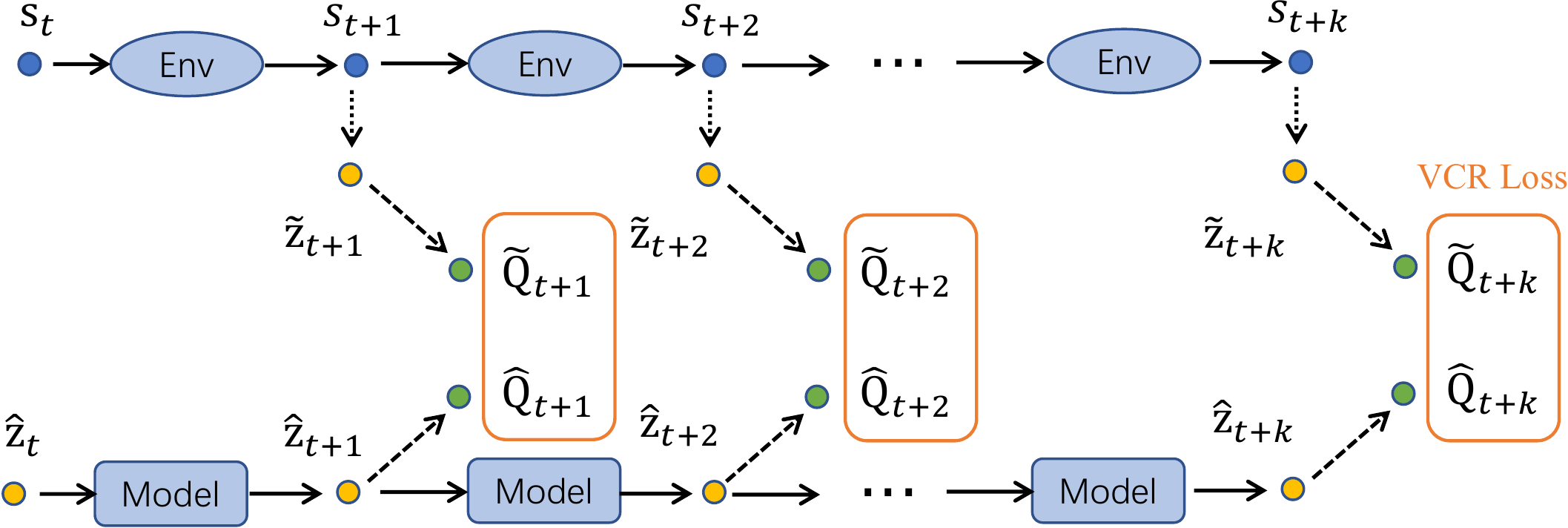}
    \caption{
    \textbf{Pipeline of \ourname Representation Learning.}  
    The agent interacts with the environment to produce a sequence of states $\{s_{t+k}\}_{k=1:K}$. The sequence is encoded into a latent embedding space, denoted as $\{\tilde{z}_{t+k}\}_{k=1:K}$. Then the parametric transition model predicts the sequence of latent states $\{\hat{z}_{t+k}\}_{k=1:K}$ conditioned on the start point $\hat{z}_{t}$ and the action sequence, where $\hat{z}_{t}$ is the latent embedding from ${s}_{t}$. \ourname Representation Learning enforces the value prediction from imagined states to be consistent with the values of a real state from the environment. SPR loss is also utilized to help stabilze the training at the early stage. For simplicity, SPR loss and action inputs to the environment and the transition model are omitted in the figure.}
    \label{fig:pipeline}
\end{figure}

\subsection{Preliminaries: Value-Based Reinforcement Learning}
Reinforcement Learning (RL) addresses the problem of sequential decision making, which is usually formulated with a Markov Decision Process (MDP). 
A typical MDP is represented with a tuple $\langle\mathcal{S}, \mathcal{A}, {T}, r, \gamma\rangle$.
Here $\mathcal{S}$ is a finite set of states; $\mathcal{A}$ is the action space; ${T}(s, a, s^\prime) = P(s^\prime|s,a)$ is the dynamics function describing the probability of transitioning from a state $s$ to $s^\prime$ after taking an action $a$; $r(s,a,s^\prime)$ and $\gamma\in (0,1]$ are the reward function and the discount factor respectively. 
The fundamental goal of RL is to learn an agent maximizing the discounted cumulative reward (\ie, return) $G_t = \sum_{\tau=0}^{\infty} \gamma^\tau r_{t+\tau+1}$ at any time step $t$.
The behavior of an agent is denoted by a policy $\pi(a|s)$ mapping from states to actions, while the action-value function $Q_\pi(s,a) = \mathbb{E}_\pi \left[G_t | s_t=s, a_t = a\right]$ predicts the expected return if the agent takes an action $a$ for state $s$ at the time step $t$, following the policy $\pi$.

There are various ways to learn the optimal policy $\pi^{*}$. In this paper, we focus on value-based RL algorithms that are rooted in $Q$-learning. $Q$-learning performs approximate dynamic programming by following the Bellman equation~\cite{sutton2018reinforcement}. 
Deep $Q$ Network (DQN)~\cite{DQN} scales $Q$-learning to large state (\textit{e.g.}, visual inputs) by utilizing neural networks to encode states and generate $Q$ values. 
Additionally, experience replay and a separated target network are used to stabilize the training of DQN. The overall objective for DQN is given by 
\begin{scriptsize}
\begin{equation}
\label{eq:dqn}
    \mathcal{J}(\theta) = \mathbb{E}_{(s,a,s^\prime,r) \sim \mathcal{D}}\left[ \left(r+\gamma \max_{a^\prime} Q_{\bar{\theta}}(s^\prime, a^\prime) - Q_\theta(s,a)\right)^2 \right],
\end{equation}
\end{scriptsize}
where $Q_\theta$ is the $Q$ network parameterized with $\theta$, $Q_{\bar{\theta}}$ is the target $Q$ network, and $\mathcal{D}$ represents the replay buffer which stores the experience tuples. Let $\bar{G}_t^{(n)} = \sum_{\tau=1}^n \gamma^{\tau-1} r_{t+\tau} + \gamma^n \max_a Q_{\bar{\theta}}(s_{t+n}, a)$ be the $n$-step value target. Note that when $n=1$, it reduces to the estimator used in Eqn.~\eqref{eq:dqn}, but $n>1$ is also commonly used to get a better estimation~\cite{sutton2018reinforcement,hessel2018rainbow}.

As the main goal of our work is to boost the sample efficiency of RL algorithms, without loss of generality, we build our method upon two best-performing algorithm variants, \textit{i.e.}, Rainbow DQN for discrete action domains and Soft Actor-Critic (SAC) for continuous action domains. 
For more details, please refer to Appendix~\ref{appendix:sac} or their papers. Though empirical results are only shown with these two algorithms, our method is general enough to be easily integrated into any other value-based RL algorithms.

\subsection{\ourname Representation Learning}

Various methods have shown that representation learning can boost the sample efficiency of RL. 
However, all these methods approach the problem from a computer vision perspective, \textit{i.e.,} encouraging similar states to generate similar embeddings and forcing different states to be discriminative.
Despite their effectiveness, visual recognition is not always directly related to decision-making. 
To alleviate this issue, we base our method on the assumption~\cite{SPR,yu2021playvirtual} that a good representation for RL should be able to predict the resulting state following a sequence of actions. Instead of aligning the predictions in the embedding space, our key idea is the value prediction from an \emph{imagined state} should be consistent with the values of a \emph{real state} from the environment, as illustrated in Fig.~\ref{fig:pipeline}. Thus, the method is termed \ourname Representation Learning (\shortnameno).



\paragraph{State Prediction with Transition Model.}
Considering a one-step interaction between an agent and the environment: $(s_t, a_t, s_{t+1})$, $s_{t+1}$ can be determined by the transition $T$ in the underlying MDP, given $(s_t, a_t)$. Similar to SPR and PlayVirtual, we introduce a parametric transition model $h(\cdot,\cdot)$ to mimic the behavior of $T$ in a latent embedding space. More specifically, a (convolutional) neural encoder $f(\cdot)$ is used to encode a pixel-based observation/state $s_t$ into a latent representation $z_t = f(s_t)$. Then $h(\cdot,\cdot)$ operates by $\hat{z}_{t+1} = h(z_t, a_t)$.   As shown in Fig.~\ref{fig:pipeline}, based on the current latent state $z_t$, following a sequence of future $K$ actions $a_{t:t+K-1}$, a sequence of state predictions $\hat{z}_{t+1:t+K}$ is obtained by applying $h(\cdot, \cdot)$ recursively:
\begin{equation}
\label{eqn:dynamic}
\begin{aligned}
        \hat{z}_t &= {z}_t = f(s_t) \\
        \hat{z}_{t+k+1} &= h(\hat{z}_{t+k}, a_{t+k}), \quad k = 0, 1, \dots, K-1.  \\
    \end{aligned}
\end{equation}
The transition model $h(\cdot, \cdot)$ is usually optimized to minimize the prediction error between $\hat{z}_{t+1:t+K}$ and $\tilde{z}_{t+1:t+K}$, a sequence of latent features extracted directly from the raw observations / states with $\tilde{z}_{t+k} = f_\text{T}(s_{t+k})$. For example, the self-predictive representations (SPR)~\cite{SPR} method utilizes a cosine similarity for prediction error, leading to the following objective:
\begin{equation}
    \mathcal{L}_{\text{SPR}} = - \sum_{k=1}^{K} \left(\frac{\hat{z}_{t+k}}{||\hat{z}_{t+k}||_2}\right)^\top \left(\frac{\tilde{z}_{t+k}}{||\tilde{z}_{t+k}||_2} \right),
    \label{spr_loss}
\end{equation}
where $\tilde{z}$ is often referred to as the target embedding and generated from a target encoder $f_\text{T}(\cdot)$, which is a stop-gradient version of the online encoder $f(\cdot)$.


\paragraph{Value Consistency.}
\label{sec:value-consistency}
Intuitively, for a good encoder and a reasonable transition model in RL algorithms, the predictive representations should contain abundant information such that precise value estimation can be made by feeding the latent embedding into a value head. 
With a slight abuse of notation, we denote the $Q$-value predictions by a value head for the embedding-action pair $(z_t, a_t)$ as $Q(z_t, a_t)$. For simplicity, we use $Q(z_t,\cdot)$ to denote the action values from the embedding $z_t$ for all possible actions. In this way, the value predictions for imagined states $\hat{z}_{t+1:t+K}$ and target states $\tilde{z}_{t+1:t+K}$ can be written as $\{Q(\hat{z}_{t+k}, \cdot)\}_{k=1:K}$ and $\{Q_{\text{T}}(\tilde{z}_{t+k}, \cdot)\}_{k=1:K}$ respectively. Then our value-consistent representation learning loss is obtained by minimizing the distance between these two value predictions, described by:
\begin{equation}
\mathcal{L}_{\text{VCR}} = \sum_{k=1}^{K} d_{\text{VCR}}\left(Q(\hat{z}_{t+k}, \cdot), Q_{\text{T}}(\tilde{z}_{t+k}, \cdot)\right),
\end{equation}
where $d_\text{VCR}$ is a distance metric for action-values to be detailed below, and $Q_{\text{T}}(\tilde{z}_{t+k}, \cdot)$ is  the target value prediction generated with a target head $Q_{\text{T}}$.
\paragraph{Overall Objective.} At the early stage of training, the value estimation is not accurate and therefore sole \shortname may not provide good supervision signals for training the dynamics model. SPR loss is introduced to help stabilize the training of the dynamics model. Now, we are ready to present our overall training objective as below:
\begin{equation}
    \mathcal{L}_{\text{total}}(\theta) = \mathcal{J}(\theta) +\lambda_{\text{SPR}}\mathcal{L}_{\text{SPR}}(\theta)+\lambda_\text{VCR}\mathcal{L}_\text{VCR} (\theta),
    \label{equ:total_loss}
\end{equation}
where $\theta$ denotes all the model parameters used for computing the above loss function, and $\lambda_\text{SPR}$, $\lambda_\text{VCR}$ are the hyperparameters to weight different losses. 


\paragraph{Value-Consistent Distance Metric.} Here we develop our distance metric $d_\text{VCR}(\cdot, \cdot)$ and provide two different implementations for both discrete action and continuous settings. It seems easy to come up with an idea for $d_\text{VCR}(\cdot, \cdot)$. For example, one can simply apply mean-squared loss to the imagined q-values and the target q-values over all possible actions: $d_\text{MSE} = \frac{1}{|\mathcal{A}|} \sum_{a\in \mathcal{A}} [Q(\hat{z}_t, a) - Q_{\text{T}}(\tilde{z}_t, a)]^2 $. Actually, this can not work through because the target action-value $Q(\tilde{z}_t, a_t)$  for the real action $a_t$ keeps evolving as the training proceeds. Based on $\mathcal{J}(\theta)$ in Eqn.~\eqref{eq:dqn}, at each iteration, the Q value $Q(\tilde{z}_t, a_t)$ (where $\tilde{z}_t = f(s_t)$) for a real state-action pair $(s_t, a_t)$ will be updated towards a $n$-step target estimation $\bar{G}_t^{(n)}$. If we still use $d_\text{MSE}$, it means that we are optimizing $Q(\hat{z_t}, a_t)$ towards a sub-optimal point. Based on the above observation, we propose the following distance function:
\begin{equation}
\label{eq:vcr_distance}
\begin{aligned}
d_\text{VCR} &= \frac{1}{|\mathcal{A}|} \sum_{a\in \mathcal{A}} \left[Q(\hat{z}_t, a) - \bar{Q}(\tilde{z}_t, a)\right]^2, \\
\text{where} & \quad \bar{Q}(\tilde{z}_t, a) = \left\{ \begin{array}{ll}
    \bar{G}_t^{(n)} & \text{if} \quad a = a_t, \\
    Q_{\text{T}}(\tilde{z}_t, a) & \text{if} \quad a \ne a_t.
\end{array} \right.
\end{aligned}
\end{equation}
With above equations, we let the value prediction of a real state-action pair $(s_t, a_t)$ align with a $n$-step target estimation $\bar{G}_t^{(n)}$, while other pairs align with the corresponding target action-value $Q_{\text{T}}(\tilde{z}_t, a)$. To avoid the trivial solution problem\cite{BYOL}, we stop the gradient of the target value  $Q_{\text{T}}(\tilde{z}_t, a)$. Additionally, the varying rewards introduced by $\bar{G}_t^{(n)}$ also disqualify the constant output as a solution.

\emph{Discrete Action Implementation.} For discrete actions, the $Q$ network or head is implemented to directly generate $|\mathcal{A}|$ outputs representing the $Q$ values for the corresponding actions. We can simply enumerate all of them to calculate the above distance in Eqn.~(\ref{eq:vcr_distance}).
The above $\bar{G}_t^{(n)}$ is given by $\bar{G}_t^{(n)} = \sum_{\tau=1}^n \gamma^{\tau-1} r_{t+\tau} + \gamma^n \max_a Q_{\bar{\theta}}(s_{t+n}, a)$, which adopt the same form with $n$-step estimation of $Q$-learning to mitigate possible gradient conflict in multi-task learning\cite{sener2018multi, yu2020gradient, jean2019adaptive}.
As our method is based on Rainbow, each value is divided into bins to build a distribution.

\emph{Continuous Action Implementation.} For continuous actions, the $Q$ network has a different implementation, which usually takes both the state $s$ and the action $a$ and then outputs a scalar as the $Q$ value $Q(s, a)$. Note that SAC is chosen to be our baseline algorithm, in which the target $Q$ value is estimated by one-step return.
For the case that action $a$ is real action $a_t$, the above $\bar{G}_t^{(n)}$ is calculated by $\bar{G}_t^{(n)} = r_{t} + \gamma Q_{\phi_{targ}}(s_{t+1}, a_{\theta}(s_{t+1}))$. 
We randomly sample a few actions as actions that are not equal to $a_t$, which together with $a_t$ constitute the set $\mathcal{A}$.
Considering soft $Q$ values are used in SAC, we also employ soft $q$ values when calculating the value-consistent loss.
\section{Experiments}
In this section, we first investigate the quality of dynamic models given by SPR in terms of value estimation, and show that \shortname is able to improve it substantially.
Further, we empirically evaluate the performance of our \shortname on boosting data efficiency in Atari 100K~\cite{bellemare2013ALE, simple} and DeepMind Control Suite \cite{tassa2018DMC}. We then conduct ablation studies to analyze the important components in our method.

\subsection{Value of Imagined States}
\label{sec:q-error}
Intuitively, it's a good property for reasonable encoder and
transition model that the predicted representations contain abundant information
such that precise value estimation can be made. We measure the absolute difference between the values of imagined state $\hat{z}_{t+k}$ and true values, where $\hat{z}_{t+k}$ is given by Eqn.~\ref{eqn:dynamic}. Here we use the Monte-Carlo return as true value. For a complete evaluation trajectory of length $T$, the average error is as:
\begin{equation}
\label{eqn:q-error}
\frac{1}{TK} \sum_{t=1}^{T} \sum_{k=1}^{K} \big | Q(\hat{z}_{t+k}, a_{t+k}) - G_{t+k}  \big | \cdot \mathbbm{1}_{(t+k \leq T)}.
\end{equation}
The timesteps that go beyond the terminal timestep are masked. Out of implementation efficiency, we collect evaluation trajectories of total length 1000 and test the error every 1000 train steps. In finite-horizon Atari games, $G_{t+k}$ is calculated as the discounted return. DeepMind Control Suite environments are infinite control problem with time limit $L=1000$, so $G_{t+k}$ is derived by $L-t-k$ step bootstrap of $Q(s_L, a_{\theta}(s_L))$. For fair comparison, we set prediction step $K$ for SPR and \shortname to be the same ($K=5$ for Atari 100K, and $K=3$ for DeepMind control Suite).

Fig.~\ref{fig:q-error} compares the average  shows the results in a subset of Atari 100K and DeepMind Control Suite. For most curves of SPR, as the policy immediately achieves high returns at the beginning steps, the $Q$ error of SPR grows rapidly and then keeps high until the end. Instead, \shortname consistently has smaller $Q$ error over x-axis in almost all environments. On ball-in-cup-catch and walker- walk, the error of \shortname reduces to a very low value at the right side of step axis. Note that in the environments where \shortname has more accurate values of predicted representations than SPR in Fig.~\ref{fig:q-error}, \shortname usually achieves a higher or comparable score than SPR (For scores of individual games, refer to Table~\ref{table:dmc_compare} and Table \ref{table:atari_compare_entry} in the appendix).

\subsection{Setup for Empirical Evaluation}
\label{sec:setup}
\paragraph{Environments.} We benchmark \shortname in environments where the number of interactions is limited.
Specifically, we choose Atari 100K for discrete control and DeepMind Control Suite for continuous control. 
For DeepMind Control Suite, following \citeauthor{hafner2019planet} \shortcite{hafner2019planet} and \citeauthor{DrQ} \shortcite{DrQ}, we use six environments (\ie, ball-in-cup, finger-spin, reacher-easy, cheetah-run, walker-walk, and cartpole-swingup) for benchmarking with 100K and 500K \emph{environment steps}.

\paragraph{Baselines.} \emph{For Atari 100K}, we take SPR as a \emph{strong baseline}. Also,  SimPLe~\cite{SimCLR}, DER~\cite{DER}, OTR~\cite{OTR}, CURL~\cite{srinivas2020curl}, DrQ~\cite{DrQ} are chosen as baselines because all of them were state-of-the-arts in Atari 100K at their publications.  PlayVirtual~\cite{yu2021playvirtual} is chosen as another state-of-the-art baseline, which is also a representation learning method based on SPR requiring a little more computation and memory. 
\emph{For DeepMind Control Suite}, we choose Dreamer~\cite{hafner2019Dreamer}, SAC+AE~\cite{yarats2019improving}, SLAC~\cite{lee2020SLAC}, CURL, DrQ, SPR and PlayVirtual as our baselines. Since SPR is designed for discrete tasks, we adopt a modified SPR based on SAC for continuous tasks\footnote{Link: \url{https://github.com/microsoft/Playvirtual}, MIT License.}. EfficientZero \cite{efficientzero}, a method based on Monte-Carlo Tree Search, has achieved excellent performance on both Atari 100K and DeepMind Control 100K. However, considering the success comes at the cost of one order of magnitude more GPU and CPU computation complexity, we do not compare with EfficientZero here.

\paragraph{Implementation Details.} 
\emph{For discrete action tasks}, we base our implementation of \shortname on the official code\footnote{Link: \url{https://github.com/mila-iqia/spr}, MIT License.} of SPR.
The encoder is a three-layer convolutional network, and the transition model is a two-layer convolutional network. $Q$ head is a two-layer linear network shared by $Q$-learning and \ourname representation learning. For $Q$ head, noisy parameters \cite{noisynet} are reserved because we verify that the noisy q-value output does not have any negative influence on \ourname representation learning (see Appendix~\ref{appendix:result}).
Different from SPR which has an asymmetric prediction head at the end of the online encoding branch, we validate that the removal of the prediction head would not impair the performance (see Appendix~\ref{appendix:result}). Thus we directly build $q$ head following the encoding branch.
The prediction step is set $K=5$. $Q$-learning loss and \ourname loss are optimized jointly by an Adam Optimizer~\cite{kingma2014adam}, where the batch size is 32. 
\emph{For continuous control tasks}, the modified SPR for continuous control is chosen as our codebase.
Prediction step is set $K=3$. Actor loss, critic loss, and \ourname loss are optimized separately by three Adam optimizers, where the batch size for the actor-critic update is 512 and the batch size for \shortname update is 128. For more details, please refer to Appendix~\ref{appendix:training}. Code will be open-sourced upon acceptance. 

\paragraph{Evaluation Metrics.}
According to ~\citeauthor{agarwal2021precipice} \shortcite{agarwal2021precipice}, we choose \emph{interquartile mean} (IQM) and \emph{optimality gap} as \emph{main evaluation indicator} considering their good properties. IQM computes the mean Human Normalized Score (HNS) of the middle 50\% runs over all games and seeds. Optimality gap denotes the gap between algorithms and the target performance. Higher IQM and lower optimality gap are better. We also present performance profile curves.

\begin{figure}[t]
    \centering
    \includegraphics[width=.95\columnwidth]{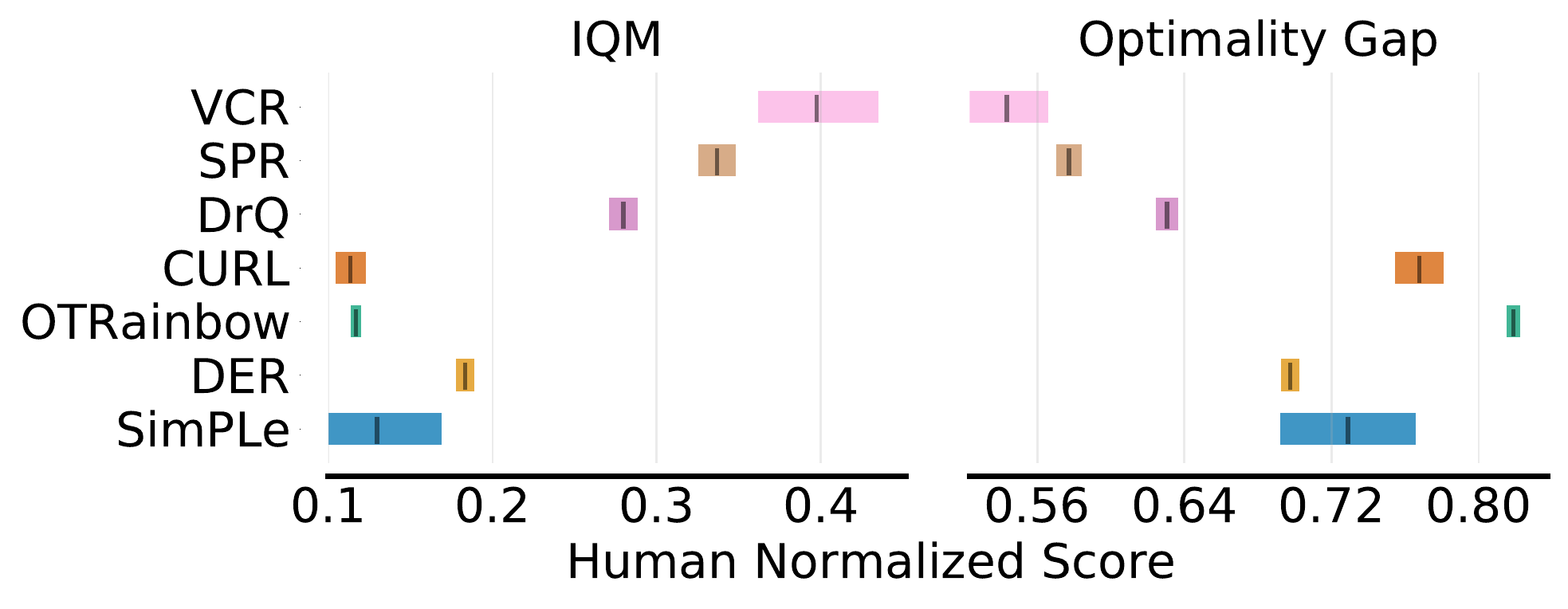}
    \caption{Aggregate IQM and optimality gap of methods. Higher IQM and lower optimality gap are better. The shaded bar shows 95\% stratified bootstrap confidence intervals \cite{agarwal2021precipice}. \shortname runs 10 seeds over 26 games.
    }
    \label{fig:atari-IQM}
\end{figure}

On Atari-100K and DeepMind Control 100K, we use 10 seeds for each game to evaluate \shortnameno.  
For Atari 100K aggregate metrics, We use rliable results with 100 random seeds from \citeauthor{agarwal2021precipice} \shortcite{agarwal2021precipice} for baselines that have been evaluated in this paper\footnote{Data can be found in a public cloud bucket at \url{https://console.cloud.google.com/storage/browser/rl-benchmark-data/atari_100k}.}. For SimPLe and PlayVirtual, we use results from its paper~\cite{simple,yu2021playvirtual}. For DeepMind Control, we use results from orginal papers except for SPR and PlayVirtual on 100K steps, in which we rerun the official code with 10 seeds to get every single run scores for calculating IQM scores. Limited by computational resource, on 500K steps, we directly report results from their paper. Without access to baselines' single run scores for calculating IQM score, we simply report median score for 500K.

\begin{table*}[t]
    \centering

        \begin{adjustbox}{width=0.67\textwidth}
        \begin{tabular}{l c c c c c c c | c}
            \toprule
            \textbf{Game} & \textbf{SimPLe} & \textbf{DER} & \textbf{OTR} & \textbf{CURL} & \textbf{DrQ} & \textbf{SPR} & \textbf{\shortname} &\textbf{PlayVirtual}\textdagger  \\
            \midrule
            IQM HNS ($\%$)          & 13.0  & 18.3  &  11.7 &  11.3 & 28.0  & 33.7   &  \textbf{39.7}  & 37.4\\
            Optimality Gap ($\%$)   & 72.9  & 69.8  & 81.9  &  76.8 & 63.1  &  57.7  &    \textbf{54.4} & 55.8  \\
            \bottomrule
        \end{tabular}
        \end{adjustbox}
        
    \caption{Aggregated scores achieved by different methods on Atari-100k. \textdagger~ denotes using virtual trajectories.}   
    \label{table:atari_compare_aggregate}
\end{table*}

\subsection{Results of Empirical Evaluation}

\begin{figure}[t]
    \centering
    \includegraphics[width=0.95\columnwidth]{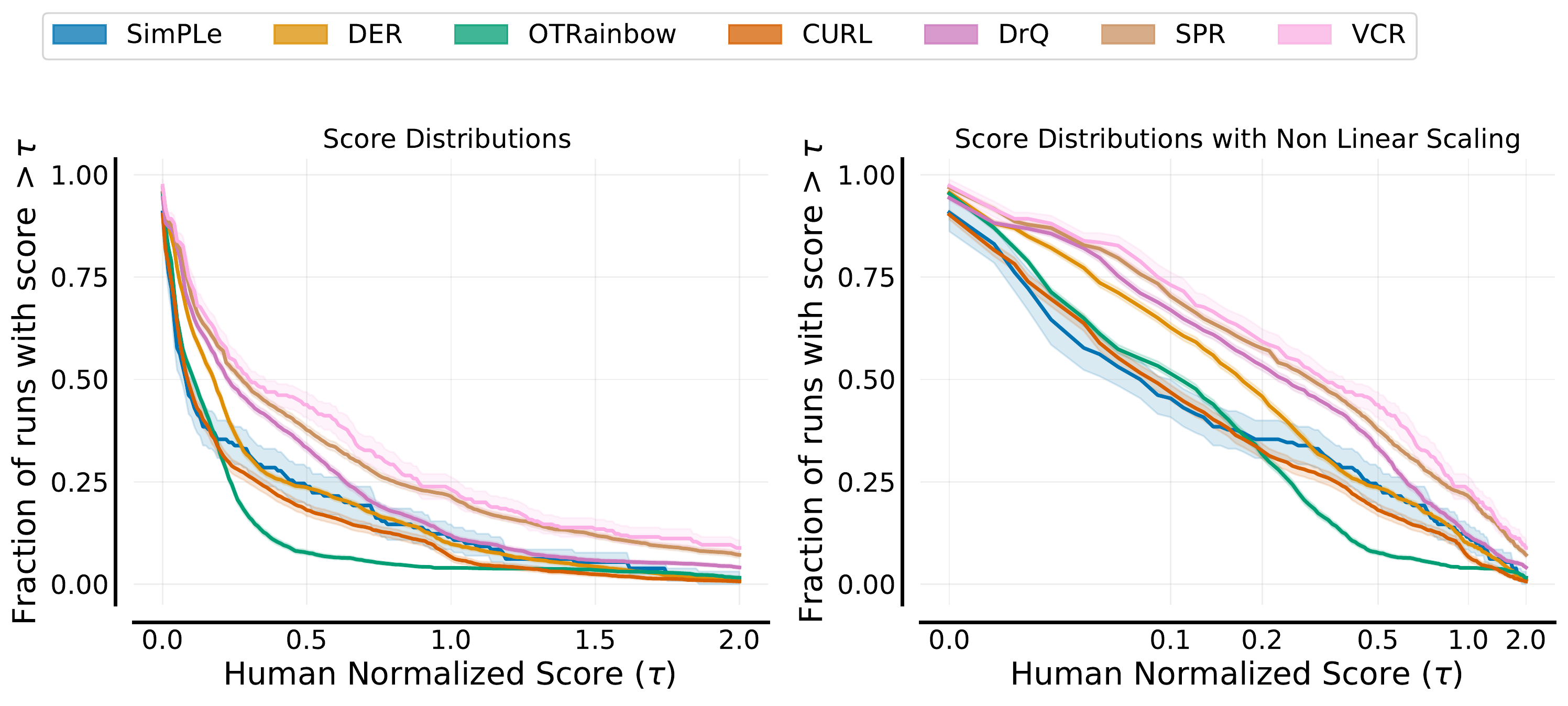}
    \caption{Performance profiles based on score distributions with linear and non-linear scaling on Atari 100K.
    }
    \label{fig:atari-curve}
\end{figure}

\paragraph{Atari-100K.} As shown in Fig.~\ref{fig:atari-IQM}, \shortname achieves the best performance on both IQM HNS and optimality gap.
It can be seen from the nonoverlapping confidence intervals that our improvement over SPR is statistically significant. 
Fig.~\ref{fig:atari-curve} reveals that \shortname is nearly above baselines along the whole axis, demonstrating consistent improvement over baselines. 
The improvement is particularly noteworthy when focusing on the HNS interval between $0.2$ and $1.0$.
The numerical performance of methods is presented in Table~\ref{table:atari_compare_aggregate}. 
By adding value consistency constraint, our method gets a boost over the baseline SPR by \textbf{6.0\% IQM HNS}, which is significant when considering the $5.7\%$ IQM HNS improvement of SPR over DrQ and $3.7\%$ of PlayVirtual over SPR. Compared to PlayVirtual, the other regularization method based on SPR architecture by producing virtual cycle trajectories, \shortname achieves better performance (higher IQM HNS by 2.3\%) with less computation and memory consumption (See Appendix~\ref{appendix:training}). All scores of individual games are shown in Table~\ref{table:atari_compare_entry}.


\begin{table*}[ht]
    \footnotesize 
    \centering
    \begin{adjustbox}{width=0.85\textwidth}
    {
        \begin{tabular}{l c c c c c c c | c}
            \toprule
            \textbf{100k Step Scores}  & \textbf{Dreamer} & \textbf{SAC+AE} & \textbf{SLAC} & \textbf{CURL} & \textbf{DrQ} & \textbf{SPR} &  \textbf{\shortname} & \textbf{PlayVirtual}\textdagger \\ \hline
            \specialrule{0em}{1.5pt}{1pt}   
            Finger, spin  & 341 $\pm$ 70   & 740 $\pm$ 64    & 693 $\pm$ 141 & 767 $\pm$ 56  & \textbf{901 $\pm$ 104}  & 840 $\pm$ 143    &  795 $\pm$ 157 & 683 $\pm$ 189 \\
            Cartpole, swingup   & 326 $\pm$ 27   & 311 $\pm$ 11    & -             & 582 $\pm$ 146 &  759 $\pm$ 92 & \textbf{815 $\pm$ 48}    & \textbf{815 $\pm$ 47} & 812 $\pm$ 66\\
            Reacher, easy   & 314 $\pm$ 155  & 274 $\pm$ 14    & -             & 538 $\pm$ 233  & 601 $\pm$ 213 & 684 $\pm$ 186    &  \textbf{763  $\pm$ 112}& 663 $\pm$ 214\\
            Cheetah, run   & 235 $\pm$ 137  & 267 $\pm$ 24    & 319 $\pm$ 56  & 299 $\pm$ 48  & 361 $\pm$ 67  & 452 $\pm$ 117    & 422  $\pm$ 54 & \textbf{510 $\pm$ 38}\\
            Walker, walk     & 277 $\pm$ 12   & 394 $\pm$ 22    & 361 $\pm$ 73  & 403 $\pm$ 24  & 634 $\pm$ 160  & 397 $\pm$ 220    & \textbf{650 $\pm$ 143} & 499 $\pm$ 161 \\
            Ball in cup, catch   & 246 $\pm$ 174  & 391 $\pm$ 82    & 512 $\pm$ 110 & 769 $\pm$ 43  & 914 $\pm$ 51  & 807 $\pm$ 165   & 858  $\pm$ 85 & \textbf{939$\pm$ 20}  \\ 
            \midrule
            IQM HNS        & - & - & - & - & 731 & 700 & \textbf{745} & 690 \\
            Optimality Gap & 710  & 603 & - & 440 & 305 & 335 & \textbf{283} & 316 \\
            \midrule
            \textbf{500k Step Scores}   \\ \midrule
            Finger, spin  & 796 $\pm$ 183  & 884 $\pm$ 128   & 673 $\pm$ 92  & 926 $\pm$ 45 & 938 $\pm$ 103 & 924 $\pm$ 132 & \textbf{972 $\pm$ 25} & 963 $\pm$ 40 \\
            Cartpole, swingup   & 762 $\pm$ 27   & 735 $\pm$ 63    & -             & 841 $\pm$ 45& 868 $\pm$ 10 & \textbf{870 $\pm$ 12}  & 854 $\pm$ 26 & 865 $\pm$ 11 \\
            Reacher, easy  & 793 $\pm$ 164  & 627 $\pm$ 58    & -             & 929 $\pm$ 44 & \textbf{942 $\pm$ 71} & 925 $\pm$ 79 & 938 $\pm$ 37 & \textbf{942 $\pm$ 66} \\
            Cheetah, run  & 570 $\pm$ 253  & 550 $\pm$ 34    & 640 $\pm$ 19  & 518 $\pm$ 28  & 660 $\pm$ 96 & 716 $\pm$ 47 & 661 $\pm$ 32 & \textbf{719 $\pm$ 51}  \\
            Walker, walk   & 897 $\pm$ 49   & 847 $\pm$ 48    & 842 $\pm$ 51  & 902 $\pm$ 43  & 921 $\pm$ 45 & 916 $\pm$ 75 & \textbf{930 $\pm$ 18} & 928 $\pm$ 30 \\
            Ball in cup, catch  & 879 $\pm$ 87   & 794 $\pm$ 58    & 852 $\pm$ 71 & 959 $\pm$ 27 & 963 $\pm$ 9 & 963 $\pm$ 8  & 958 $\pm$ 4 & \textbf{967 $\pm$ 5} \\
             \midrule
            Median Score  & 794.5 & 764.5 & 757.5 & 914.0 & 929.5 & 920.0 & \textbf{934.0}  & \textbf{935.0} \\
            \bottomrule
        \end{tabular}
    }
     \end{adjustbox}
    \caption{Scores (mean and standard deviation) achieved by different methods on the DeepMind Control. We run \shortname with 10 seeds. On 500K steps, single run scores of SPR and PlayVirtual are missing to calculate IQM, so we follow \citeauthor{yu2021playvirtual} \shortcite{yu2021playvirtual} to report median scores to profile the overall performance. \textdagger~ denotes using virtual trajectories.
    }
    \label{table:dmc_compare}
\end{table*}

\paragraph{DeepMind Control Suite.} Within very limited interactions 100K, \shortname achieves the best performance on 3 out of 6 tasks, as shown in  Table~\ref{table:dmc_compare}. In addition, \shortname achieves the best IQM HNS and optimality gap. Compared with the baseline SPR, \shortname has a relatively 6.4\% higher IQM and 15.5\% lower optimality gap. When 500K ineractions are allowed, \shortname is close to the perfect score in 4 environments, and achieves a comparable median score with DrQ, SPR and PlayVirtual.

\subsection{Analysis}
\label{sec:ablation}

In this section, we conduct ablation studies with Atari 100K to analyze the important components in \shortname and the effectiveness of \shortname compared with other auxiliary losses. 

\paragraph{Value-Consistent Distance Metric.}We find that the Value-Consistent distance metric has a great influence on the performance. We use a mixed target of $n$-step target estimation $\bar{G}_t^{(n)}$ and action-value $Q_{\text{T}}(\tilde{z}_t, a)$, \ie, $d_\text{VCR}$ (see Eqn.~\ref{eq:vcr_distance}). Here, we test two variants to validate the proposed Value-Consistent distance metric. The first is to simply apply MSE loss to the imagined q-values and the target q-values over real state-action pairs: $d_\text{MSE} = [Q(\hat{z}_t, a_{t}) - Q_{\text{T}}(\tilde{z}_t, a_{t})]^2 $.
Further, we can enforce value consistency over all possible actions: $d_\text{MSE-A} = \frac{1}{|\mathcal{A}|} \sum_{a\in \mathcal{A}} [Q(\hat{z}_t, a) - Q_{\text{T}}(\tilde{z}_t, a)]^2 $. 
Note that MSE loss is replaced with a cross-entropy for distributional RL.

We evaluate these two variants on Atari 100K. 
$d_\text{MSE}$ achieves $30.9\%$ IQM HNS and $60\%$ optimality gap, while $d_\text{MSE-A}$ achieves $34.0\%$ IQM HNS and $57.9\%$ optimality gap.
Compared to $d_\text{VCR}$ ($39.7\%$ IQM HNS and $54.4\%$ optimality gap), two variants $d_\text{MSE}$ and $d_\text{MSE-A}$ have inferior performance.
Especially for $d_\text{MSE}$, it is lower than the baseline ($33.7\%$ IQM HNS and $57.7\%$ optimality gap). 
That may be because, in these two variants, \shortname loss has conflicting gradients with policy learning loss (\ie, DQN loss), a common issue in multi-task learning~\cite{sener2018multi, yu2020gradient, jean2019adaptive}. 
$Q$-learning loss pushes the value function towards the value distribution under the optimal policy while $d_\text{MSE}$ and $d_\text{MSE-A}$ aim to align with the current value approximation. 
Intuitively, $d_\text{MSE}$ has bigger conflict than $d_\text{MSE-A}$ because only real state-action pairs lead to an explicit conflict, which are only $1/|\mathcal{A}|$ of state-action pairs used in $d_\text{MSE-A}$, where  $|\mathcal{A}|$ is the size of the action set. It may explain why $d_\text{MSE}$ has a larger drop in performance. 

\paragraph{Comparison with Reward Loss.} One may simply attribute the improvement of \shortname to the introduction of reward to representation learning in $\bar{G}_t^{(n)}$. Here, we construct a baseline by adding reward prediction based on SPR, where the dynamics model outputs predicted reward and next state conditioned on current state and action. The predicted reward is supervised by real reward. We denote this baseline as \emph{SPR with reward loss}, which achieves $35.4\%$ IQM HNS and $56.8\%$ optimality gap. 
We see that although \emph{SPR with reward} achieves a little higher IQM HNS than baseline, it is still far behind \shortnameno. This implies that the effectiveness of our method should be attributed to consistent value prediction rather than involving additional reward prediction.

\paragraph{Comparison with Policy Learning Loss.} \ourname loss and policy learning loss are similar in terms of the formula form, both of which leverage $Q$-values to update the encoder and $Q$-value head. 
 For example, in the discrete setting, both  $\mathcal{L}_\text{VCR}$ and $\mathcal{L}_\text{DQN}$ align predicted $Q$-values to $n$-step estimation, while $\mathcal{L}_\text{VCR}$ is computed over $K$ times of state-action pairs, where $K$ is prediction steps. Thus a possible concern is that the boost of our method comes from more state-action pairs to update the $Q$-value head. Here we replace $\mathcal{L}_\text{VCR}$ on imagined state-action pairs with $\mathcal{L}_\text{DQN}$, equivalent to increasing the mini-batch size of $\mathcal{L}_\text{DQN}$. 
\emph{SPR-L} achieves $27.2\%$ IQM HNS and $61.5\%$ optimality gap, while \emph{SPR-XL} achieves $17.6\%$ IQM HNS and $70.1\%$ optimality gap.
Two variants display performance drop, which implies \shortname helps train the $Q$-value head but the main gain comes from its regularization on representation learning of the encoder and transition model.

\paragraph{Influence of Prediction Steps K.} We increase the number of prediction steps from 5 to 9 to test if more improvement can be obtained. 
$K=9$ achieves $39.7\%$ IQM HNS and $54.3\%$ optimality gap, which is roughly comparable to $K=5$. That means increasing prediction steps in a range would not change the overall performance although the performance on a subset of games increases much (see Appendix~\ref{appendix:result}), at the cost of more computation and memory. 



\vspace{-5pt}
\section{Conclusion and Limitation}
\label{conclusion}
To boost the sample efficiency of value-based reinforcement learning algorithms, we propose a novel \emph{Value-Consistent Representation Learning} (VCR) method. The intuition behind VCR is that an agent should be capable of making imagination of future states from its behaviors and obtaining correct value prediction based on the imagined states. The property becomes more demanding when the environment is stochastic and learning a precise transition model is impossible. Some previous works have validated the effectiveness of this idea with policy-based methods. We develop a value-consistent metric for $Q$ values and introduce it into value-based RL algorithms for the first time. More specifically, given a real trajectory of interactions with the environment, we employ a transition model in the latent space to shoot forward following the action sequence. Then we obtain $Q$-value predictions for each of the real-imagined state pairs and align them with the metric. We further show that the method is compatible with any value-based methods by providing two implementations dealing with both discrete and continuous actions. We evaluate our method on two benchmarks including Atari 100K for discrete control and DeepMind Control 100K for continuous control. The results clearly show that VCR can improve sample efficiency significantly and achieve new state-of-the-art on both tasks. However, there are still some limitations in our method. For example, we derive the value-consistent distance metric simply employing an MSE loss, which might not be robust to value scales. We leave better distance metric investigation as future work. 



\bibliography{neurips_2022}
\bibliographystyle{plain}

\clearpage

\appendix
\section*{Appendix}

\section{Algorithm}
\subsection{Soft Actor Critic}
\label{appendix:sac}

Soft Actor Critic (SAC)~\cite{haarnoja2018sac} is a widely used off-policy algorithm for continuous control, employing policy entropy regularization as part of the reward to encourage exploration. We denote the parameters of a stochastic policy $\pi$ and $Q$-function $Q$ as $\theta$ and $\phi$ separately. SAC learns a stochastic policy $\pi_\theta$, two $Q$-functions $Q_{\phi_1}$, $Q_{\phi_2}$, and a temperature weight $\alpha$, which balances exploration and exploitation. 

\paragraph{Learning Critic.} To mitigate the over-estimation problem, SAC uses the double Q-networks and takes the minimal $Q$-value between two $Q$-approximators. 
SAC sets up the loss function for critic update over transitions $(s,a,s',r, d)$ sampled from relay buffer:
\begin{equation}
    \mathcal{L}_{\phi}^{\text{critic}}=\sum_{i=1,2}{\big(Q_{\phi_i}(s,a)-y(r,s',d)\big)^2},
    \label{sac_critic_loss}
\end{equation}
where $r$ is reward, $s$ is the current state, $s'$ is the next state, $d$ is the terminal flag which equals $1$ if $s'$ is the terminal state, and else $0$. The target is cut off gradient, given by
\begin{equation}
    y(r,s',d)=r+{\gamma}(1-d)\big(\min_{i=1,2}Q_{\phi_{\text{targ},i}}(s',a')-{\alpha}\log {\pi}_{\theta}(a'|s')\big),
    \label{sac_y_target}
\end{equation}
where $\gamma$ is the discounted factor, $Q_{\phi_{\text{targ},i}}$ is target Q-networks updated by an exponential  moving average (EMA) over $Q_{\phi_{i}}$.

\paragraph{Learning Actor.} The actor $\pi_\theta$ aims to take the optimal action to maximize the action value plus the policy entropy. Thus, the actor update loss is
\begin{equation}
    \mathcal{L}_{\theta}^{\text{actor}}=-\big(\min_{i=1,2}{Q_{\phi_i}(s,a_\theta(s))}-{\alpha}\log {\pi}_{\theta}(a_\theta(s)|s)\big),
    \label{eq:sac_critic_loss}
\end{equation}
where $a_\theta(s)$ is sampled from a stochastic policy $\pi_\theta(s)$.

\subsection{Pseudo Code}
Here we give the pseudo code Algo.~\ref{algo:VCR_loss} for \ourname representation learning, which can be easily integrated into value-based algorithms. For simplicity, the pseudo code ignores many details.
\setlength{\textfloatsep}{0pt}
\begin{algorithm}[tb]
\caption{\ourname Representation Learning}
\label{algo:VCR_loss}
\SetAlgoLined
     Denote parameters of the convolutional encoder $f$, transition model $h$, value head $Q$ as $\theta$\;
     Denote parameters of the target encoder $f_\text{T}$, target value head $Q_\text{T}$ as $\theta_\text{T}$\;
     Denote prediction step as $K$, \shortname batch size as $N$\;
     Denote $Q$ value loss as $L$, Denote image augmentation as \text{transform}\;
    \KwIn{a minibatch of sequences of $(s_0,a_0,r_0,..,s_K,a_K,r_K)$\ sampled from replay buffer $\mathcal{D}$;}
    
    \For{k in (1, \ldots, K)}{
        $s_k \gets \text{transform}(s_k)$ \tcp*{Augment the input image batch}
    }
    $\hat{z}_0 \gets f(s_0)$\;
    $l \gets 0$\;
    \For{k in (1, \ldots, K)}{
        $\hat{z}_k \gets h(\hat{z}_{k-1}, a_{k-1})$\;
        $\Tilde{z}_k \gets f_\text{T}(s_{k})$\;
        $\mathcal{L}_{\text{SPR}} = -\left(\frac{\hat{z}_{k}}{||\hat{z}_{k}||_2}\right)^\top \left(\frac{\tilde{z}_{k}}{||\tilde{z}_{k}||_2} \right)$\;
        sample an action set $\mathcal{A}$ that do not include $a_k$\;
        calculate $\bar{G}_k^{(n)}$ according to the specific RL algorithm\;
        $\mathcal{L}_\text{VCR} \gets L(Q(\hat{z}_k, a_k), \bar{G}_k^{(n)})$   \tcp*{$\mathcal{L}_\text{VCR}$ for real action}
        \For{$a$ in $\mathcal{A}$}{
            $\mathcal{L}_\text{VCR} \gets \mathcal{L}_\text{VCR} +  L(Q(\hat{z}_k, a), Q_\text{T}(\tilde{z}_k, a))$ \tcp*{$\mathcal{L}_\text{VCR}$ for other possible actions}
        }
        $l \gets l+\lambda_{\text{SPR}}\mathcal{L}_{\text{SPR}}+\lambda_\text{VCR}\mathcal{L}_\text{VCR}$\;
    }
$\theta \gets \text{optimize}(\theta, l)$\;
update $\theta_\text{T}$ by exponential moving average of $\theta$\;

\end{algorithm}

\setlength{\textfloatsep}{0pt}

\section{Experiment}
\subsection{Training Details}
\label{appendix:training}

\paragraph{Network Architecture.} \emph{For discrete action tasks}, the encoder is a three-layer convolutional network with (32, 64, 64) channels, (8,4,3) filter size and (4,2,1) strides. State embedding and action are concatenated together along the channel dimension as the input to the transition model, which is a two-layer convolutional network with 64, 64 channels. 
$Q$-head is a two-layer linear network shared by $Q$-learning and \ourname representation learning with hidden unit size 256. Noisy parameters, dueling network, and double $Q$ are utilized in $Q$ head. Following SPR~\cite{SPR}, the first layer of $Q$ head is reused as projection head for SPR loss with noisy parameters and dueling network turned off. A prediction head is leveraged to calculate SPR loss, but not used for \shortname loss. The target encoder $f_\text{T}$ and target $Q$-head $Q_\text{T}$ are the copy of encoder $f$ and $Q$-head $Q$ with stopping gradient.

\emph{For continuous control tasks}, following CURL~\cite{srinivas2020curl}, SPR~\cite{SPR} and PlayVirtual~\cite{yu2021playvirtual}, the encoder is a neural network with four convolutional layers and one linear layer, which outputs a 50-dimension hidden vector. The transition model is a two-layer linear network, and the first linear layer is followed by a layer normalization. Projection head and prediction head for SPR loss are built by two linear layers with 1,024 hidden units. Actor and critic are both built as a three-layer linear network, which takes as input the embedding states that the encoder produces. Exponential moving average (EMA) is employed to update the target encoder and target critic.  

\paragraph{Loss Optimization.} We split the \shortname loss in Eqn.~\ref{eq:vcr_distance} into two parts. When $a=a_t$, we denote the loss of this part as $\mathcal{L}_{\text{VCR1}}$, otherwise denote as $\mathcal{L}_{\text{VCR2}}$.
We denote the weight of $\mathcal{L}_{\text{VCR1}}$ as $\lambda_\text{VCR}$. Empirically, $\mathcal{L}_{\text{VCR2}}$ brings relatively smaller boost to performance, so a smaller weight for $\mathcal{L}_{\text{VCR2}}$ is chosen (\eg~one-tenth of $\lambda_\text{VCR}$).
\emph{For discrete control tasks}, $\lambda_\text{VCR}=0.2$ is chosen for $\mathcal{L}_{\text{VCR}}$.
Both $\mathcal{L}_{\text{VCR}}$ and $\mathcal{L}_{\text{DQN}}$ are cross entropy loss between two $Q$-value distributions. All losses are optimized jointly by an Adam optimizer.  \emph{For continuous control tasks}, $\mathcal{L}_{\text{VCR}}$ is MSE loss between two $Q$-value scalars.
$\lambda_\text{VCR}=1.0$ is set.
Actor loss, critic loss, temperature weight $\alpha$, and \shortname loss are optimized separately by four optimizers.
For both discrete and continuous control, we utilize SPR loss and a warmup for  $\lambda_\text{VCR}$ to help stabilize the training of the dynamics model at the early stage of training.
Following \cite{yu2021playvirtual}, a Gaussian ramp-up curve is used to increase $\lambda_\text{VCR}$ from 0 to the maximum at the first 50K steps, and then we keep the maximum until the end of training.
A full list of hyperparameters is provided in Table~\ref{table:atari_hyperparam} and Table~\ref{table:dmc_hyperparam}.
\begin{table}
    \centering

        \begin{tabular}{l c l}
            \toprule
            \textbf{Hyperparameter} & & \textbf{Setting} \\
            \midrule
            Gray-scaling & & True \\
            Frame stack & & 4 \\
            Observation downsampling &  & (84, 84) \\
            Augmentation  & & Random shift $\&$ \\& &intensity \\
            Action repeat &  & 4 \\
            
            Training steps &  & 100K \\
            Max frames per episode &  & 108K \\
            Evaluation trajectories & & 100 \\
            Reply buffer size & & 100K \\
            Minimum replay size for sampling & & 2000 \\
            Mini-batch size &  & 32 \\

            Optimizer &  & Adam \\
            Optimizer: learning rate & & 0.0001 \\
            Max gradient norm & & 10 \\
            Discount factor &  & 0.99 \\
            Reward clipping Frame stack &  & [-1, 1] \\
            Double Q  & & True \\
            Dueling  & & True \\
            Support of Q-distribution & & 51 bins \\
            Priority exponent & & 0.5 \\
            Priority correction  & & 0.4 $\rightarrow$ 1 \\
            Exploration & &  Noisy Net \\
            Noisy nets parameter & & 0.5 \\
            Multi-step return length &  & 10 \\
            
            Replay period every & &  1 step \\
            Number of updates per step & & 2 \\
            
            
            Target $Q$ network update period & & 1 \\
            EMA $\tau$ for target encoder & & 0 \\
            Prediction step K && 5 \\
            $\lambda_\text{SPR}$ & & 1.0 \\
            \midrule
            warmup of $\lambda_\text{VCR}$ && True\\
            $\lambda_\text{VCR}$ && 0.2 \\
            \bottomrule
        \end{tabular}
\caption{Hyperparameters for Atari. \shortnameno~-specific hyperparameters are placed in a separate column.}
\label{table:atari_hyperparam}
\end{table}

\paragraph{Evaluation Protocol.} For each run, we evaluate the agent at the end of training over $N$ complete episodes and get the average of $N$ scores (\ie, for Atari 100K, $N=100$; for DeepMind Control 100K, $N=10$).

\paragraph{Time Complexity.} Since the \ourname representation learning module is disabled during collecting trajectories, the inference time of \shortname is as much as SPR and PlayVirtual. We measure the time cost for training \shortnameno, SPR, and PlayVirtual in a cluster node with one NVIDIA A100 GPU and 16 CPU cores. Prediction step is set for the best performance. Specifically, on Atari 100K, $K=5$ for SPR and \shortnameno, $K=9$ for PlayVirtual; on DeepMind Control 100K, $K=3$ for \shortnameno, $K=6$ for SPR and PlayVirtual. The average training time of different environments is demonstrated in Table~\ref{table:time}.
\begin{table}[h]
\footnotesize 
\centering

        \begin{tabular}{l c c c c}
            \toprule
        & \textbf{SPR}  & \textbf{PlayVirtual} & \textbf{\shortname} \\
            \midrule
    Atari 100K & 5.6h  &  9.1h   &   6.4h    \\
    DMControl 100K & 5.7h  &  7.3h   &   3.1h \\

            \bottomrule
        \end{tabular}
\caption{Training time of baselines and \shortname}
\end{table}
\label{table:time}
We can see when prediction step is identical on Atari 100K, \shortname only takes 0.8 hour more than SPR. On DeepMind Control 100K, \shortname takes nearly 46\% less time because of smaller prediction step for the best setting. Additionally, we find that when prediction step is set to 3, SPR takes 2.6 hours on DeepMind Control 100K, showing \shortname increases little training time (\ie~0.5 hour). When compared with PlayVirtual, which produces tenfold virtual trajectories and takes a longer prediction step, \shortname is more friendly to computation resource.

\begin{table}[ht]
    \centering
    
        \begin{tabular}{l  l}
            \toprule
            \textbf{Hyperparameter} & \textbf{Setting} \\
            \midrule
            Frame stack &  3 \\
            Observation downsampling &(84, 84) \\
            Augmentation  & Random crop $\&$ \\ & intensity \\
            Initial exploration steps & 1000 \\
            Action repeat & 2 finger-spin, \\ & walker-walk;\\
             &  8 cartpole-swingup; \\
             & 4 otherwise \\
            Evaluation trajectories & 10 \\
            Replay buffer size & 100K \\
            Discount factor &  0.99 \\
            
            Initial temperature &  0.1 \\
            
            SAC batch size & 512 \\
            
            Actor update freq & 2 \\
            EMA  $\tau$ for target critic &  0.01 \\
            Critic target update freq & 2 \\
            EMA  $\tau$ for target encoder & 0.05 \\
            Prediction step K & 3 \\
            $\lambda_\text{SPR}$ & 1.0 \\
            
            \midrule
            Actor  \& Critic  \& Encoder opt  &  \\
            Optimizer type & Adam \\
            $(\beta_1, \beta_2)$ & (0.9, 0.999) \\
            Learning rate & 0.0002 cheetah-run \\
             &  0.001 otherwise \\
             
            \midrule
            Temperature ($\alpha$) opt &\\
            Optimizer type & Adam \\
            $(\beta_1, \beta_2)$ & (0.5, 0.999) \\
            Learning rate & 0.0002 cheetah-run \\
             &  0.001 otherwise \\
            
            \midrule
            \shortname batch size & 128 \\
            warmup of $\lambda_\text{VCR}$ & True\\
            $\lambda_\text{VCR}$ & 1.0 \\
            
            \bottomrule
        \end{tabular}
\caption{Hyperparameters for DMControl.}
\label{table:dmc_hyperparam}
\end{table}


\subsection{More Results}

\label{appendix:result}
\paragraph{Full results on Atari-100K.} In Table~\ref{table:atari_compare_entry}, we provide the performance of baselines and \shortname on individual games. \shortname achieves the best on 6 out of 26 games.

\paragraph{Influence of increasing prediction step.} When increasing prediction step $K$ from 5 to 9, the overall performance of \shortname is nearly unchanged. However, as Table~\ref{table:VCR-K9} shows, larger prediction step has greater influence on scores of a subset of games. For example, Breakout, Demon Attack, and Qbert acquire at least 50\% boost. However, Bank Heist, Crazy Climber, and Boxing suffer significant performance drop. This reveals that a longer prediction step can be applied to some specific environments to get better scores.

\paragraph{Influence of weight $\lambda_\text{\shortname}$.} We conduct experiments to study the influence of $\lambda_\text{\shortname}$. Results in Table~\ref{table:ablation_weight} show that the optimal setting is $\lambda_\text{\shortname}=0.2$ for Atari while $\lambda_\text{\shortname}=1.0$ for DeepMind Control. This may be because on Atari 100K the distributional value is used, which provides stronger supervision than a scalar.
So a smaller weight would be better.

\begin{table}[h]
\footnotesize 
\centering

        \begin{tabular}{l l c c c}
            \toprule
             & $\lambda_\text{\shortname}$ & \textbf{$0$} & \textbf{$0.2$} & \textbf{$1.0$} \\
            \midrule
            {Atari 100k}  & IQM HNS & 33.7 & \textbf{39.7} & 28.8 \\
            & optimality gap & 57.7 & \textbf{54.4} & 61.3 \\
            \midrule
            {DMControl 100K}  & IQM HNS & 700 & 707 & \textbf{740} \\
            & optimality gap & 335 & 324 & \textbf{298}\\
            \bottomrule
        \end{tabular}
\caption{The influence of weight $\lambda_\text{\shortname}$ on \shortnameno.}
\label{table:ablation_weight}
\end{table} 

\paragraph{Removal of the prediction head.} In contrastive learning, prediction head, which only follows the online branch, introduces asymmetry into the network architecture to avoid collapsed solutions~\cite{BYOL}.
But the prediction head collides with \ourname representation learning because the prediction head would project $q$-value into a latent space that does not represent RL semantics. Here we verify that removing prediction head in SPR would not impair the performance.
Specifically, based on the SPR code, we simply remove the prediction head and calculate SPR loss directly on the output of the projection head. We randomly choose three games in Atari 100K and run them with 10 seeds. HNS is shown in Table~\ref{table:no_prediction}. Considering acceptable variance, removal of the prediction head has no impact on SPR performance. It may be because the contrastive learning framework in RL setting incorporates a dynamics model, which already causes asymmetry between the online and target branch.
The conclusion provides a good basis for directly constructing \shortname loss on $q$-value without a prediction head.

\begin{table}[h]
\footnotesize 
\centering

        \begin{tabular}{l c c}
            \toprule
    
& SPR  & SPR w.o. head \\
    \midrule
\textbf{Crazy Climber} & 1.04 & 0.81          \\ 
\textbf{Pong}          & 0.43 & 0.49          \\ 
\textbf{Battle Zone}   & 0.36 & 0.35 \\

            \bottomrule
        \end{tabular}

\caption{The influence of removing prediction head on SPR.}
\label{table:no_prediction}
\end{table}


\paragraph{Influence of Noisy Net.} Rainbow DQN employs Noisy Net~\cite{noisynet} to encourage state-conditional exploration instead of $\epsilon$-greedy exploration.
When \shortname shares $Q$-head with Rainbow DQN, intuitively noisy net may cause a disadvantage to \ourname representation learning which aims to align two distributions of action values. Here we test the influence of noisy net on \shortnameno. Specifically, we train a modified \shortname implementation on 4 randomly chosen games with 10 seeds. The modified \shortname turns off noisy layers (\ie~only using parameters that represent mean) when calculating \shortname loss, but turns on noisy layers when calculating DQN loss. As Table~\ref{table:noisynet} shows, removing noisy net for \shortname loss does not boost the performance of the \shortname algorithm. So it is reasonable to share noisy $Q$-head between DQN and \shortname. It also reveals \shortnameno's robustness when some noise is added to values.
\begin{table}[h]
\footnotesize 
\vspace{2mm}
\centering


 \begin{tabular}{l  c  c}
    \toprule
     & \shortnameno & \shortname no noise \\ 
    \midrule
\textbf{Crazy Climber} & 1.24                        & 0.85     \\ 
\textbf{Pong}          & 0.61                        & 0.61     \\ 
\textbf{Battle Zone}   & 0.31                        & 0.28      \\ 
\textbf{Demon Attack}  & 0.21                        & 0.26       \\     

            \bottomrule
\end{tabular}

\caption{The influence of NoisyNet on \shortnameno.}
\label{table:noisynet}
\end{table}

\begin{table*}[t]
\footnotesize 
\centering

        \begin{adjustbox}{width=0.9\textwidth}
        \begin{tabular}{l c c l c c l c c}
            \toprule
        \textbf{Game}  & \shortnameno-K5 & \shortnameno-K9 & \textbf{Game}  & \shortnameno-K5 & \shortnameno-K9 & \textbf{Game}  & \shortnameno-K5 & \shortnameno-K9 \\
            \midrule
            Alien & 822.4 &   913.7             & Crazy Climber & 40048.4 &  23218.2          & Kung Fu Master & 19679.7  &  13229.0  \\
            Amidar & 170.6 &  160.9            & Demon Attack & 560.4 & 1178.9             & Ms Pacman & 1477.0 & 1269.1 \\
            Assault & 571.6 & 712.7              & Freeway & 18.7 & 21.7                    & Pong & 0.9 & -3.0 \\
            Asterix & 1071.5 & 939.6             & Frostbite & 2294.7 & 1897.0              & Private Eye & 98.9 & 100.0 \\
            Bank Heist & 303.7  &  190.8       & Gopher & 539.7 &   656.9                    & Qbert & 791.1 & 3376.1 \\
            Battle Zone & 13261.0 &  12357.0      &  Hero & 5838.8 & 7340.4               & Road Runner & 10746.1 & 12017.9 \\
            Boxing & 42.5 &   24.2              &  Jamesbond & 382.5 &  379.8                & Seaquest & 521.2 & 526.7 \\
            Breakout & 18.4 &   27.7              & Kangaroo & 3393.1 & 3829.7               & Up N Down & 14674.1 & 11230.2 \\
            Chopper Command & 1024.2 & 1039.5     & Krull & 4199.2 & 3987.0 &&& \\
            \bottomrule
        \end{tabular}
        \end{adjustbox}

\caption{\shortname performance on Atari 100K when prediction step $K$ is set to 5 and 9.} 
\label{table:VCR-K9}
\end{table*}

\section{Potential Negative Societal Impact}
In the paper, we develop a representation learning method that aims to accelerate the training of agents with fewer interaction steps. From this perspective, any negative societal impact that our method may cause is similar to that of general RL algorithms. We advocate that RL-based robotics systems, game AI and other applications should follow fair and safe principles.

\begin{table*}[ht]

    \footnotesize 
    \centering

            \begin{adjustbox}{width=\textwidth}
        \begin{tabular}{l c c c c c c c c c | c}
            \toprule
            \textbf{Game} & \textbf{Human} & \textbf{Random} & \textbf{SimPLe} & \textbf{DER} & \textbf{OTR} & \textbf{CURL} & \textbf{DrQ} & \textbf{SPR}  & \textbf{\shortname} & \textbf{PlayVirtual}\textdagger \\
            \midrule
            Alien           & 7127.7    & 227.8     & 616.9     & 802.3     & 570.8     & 711.0     & 865.2     & 841.9      &  822.4 &  \textbf{947.8}\\
            Amidar          & 1719.5    & 5.8       & 74.2      & 125.9     & 77.7     & 113.7     & 137.8     & \textbf{179.7}      & 170.6 &  165.3\\
            Assault         & 742.0     & 222.4     & 527.2     & 561.5     & 330.9     & 500.9     & 579.6     & 565.6      & 571.6 &  \textbf{702.3} \\
            Asterix         & 8503.3    & 210.0     & \textbf{1128.3}    & 535.4     & 334.7     & 567.2     & 763.6     & 962.5      & 1071.5 &  933.3\\
            Bank Heist      & 753.1     & 14.2      & 34.2      &  185.5     & 55.0     & 65.3     & 232.9     & \textbf{345.4}       & 303.7 & 245.9\\
            Battle Zone     & 37187.5   & 2360.0    & 4031.2    & 8977.0     & 5139.4     & 8997.8     & 10165.3     & \textbf{14834.1}      & 13261.0 &  13260.0\\
            Boxing          & 12.1      & 0.1       & 7.8       & -0.3     & 1.6     & 0.9     & 9.0     & 35.7        &  \textbf{42.5} &  38.3 \\
            Breakout        & 30.5      & 1.7       & 16.4      & 9.2     & 8.1     & 2.6     & 19.8     & 19.6     & 18.4  & \textbf{20.6}\\
            Chopper Command & 7387.8    & 811.0     & 979.4    & 925.9     & 813.3     & 783.5     & 844.6     & 946.3      & \textbf{1024.2} &  922.4\\
            Crazy Climber   & 35829.4   & 10780.5   & \textbf{62583.6}   & 34508.6     & 14999.3     & 9154.4     & 21539.0     & 36700.5      & 40048.4 &  23176.7\\
            Demon Attack    & 1971.0    & 152.1     & 208.1     & 627.6     & 681.6     & 646.5     & \textbf{1321.6}     & 517.6      & 560.4 &  1131.7 \\
            Freeway         & 29.6      & 0.0       & 16.7      &  20.9     & 11.5     & \textbf{28.3}     & 20.3     & 19.3      & 18.7 &  16.1\\
            Frostbite       & 4334.7    & 65.2      & 236.8     & 871.0     & 224.9     & 1226.5     & 1014.2     & 1170.7      & \textbf{2294.7} &  1984.7 \\
            Gopher          & 2412.5    & 257.6     & 596.8     & 467.0     & 539.4     & 400.9     & 621.6     & 660.6      &539.7 &  \textbf{684.3}\\
            Hero            & 30826.4   & 1027.0    & 2656.6    & 6226.0     & 5956.5     & 4987.7     & 4167.9     & 5858.6      & 5838.8 &  \textbf{8597.5}\\
            Jamesbond       & 302.8     & 29.0      & 100.5     & 275.7     & 88.0     & 331.0     & 349.1     & 366.5      & 382.5 &  \textbf{394.7} \\
            Kangaroo        & 3035.0    & 52.0      & 51.3     & 581.7     & 348.5     & 740.2     & 1088.4     & \textbf{3617.4}   & 3393.1 &  2384.7\\
            Krull           & 2665.5    & 1598.0    & 2204.8    & 3256.9     & 3655.9     & 3049.2     & \textbf{4402.1}     & 3681.6      & 4199.2 &  3880.7 \\
            Kung Fu Master  & 22736.3   & 258.5     & 14862.5   & 6580.1     & 6659.6     & 8155.6     & 11467.4     & 14783.2      & \textbf{19679.7} &  14259.0 \\
            Ms Pacman       & 6951.6    & 307.3     & \textbf{1480.0}  &  1187.4     & 908.0     & 1064.0     & 1218.1     & 1318.4     & 1477.0  & 1335.4 \\
            Pong            & 14.6      & -20.7     & \textbf{12.8}      & -9.7     & -2.5     & -18.5     & -9.1     & -5.4      & 0.9 & -3.0 \\
            Private Eye     & 69571.3   & 24.9      & 34.9      & 72.8     & 59.6     & 81.9     & 3.5     & 86.0     & \textbf{98.9}  & 93.9 \\
            Qbert           & 13455.0   & 163.9     & 1288.8    &  1773.5     & 552.5     & 727.0     & 1810.7     & 866.3        & 791.1  & \textbf{3620.1} \\
            Road Runner     & 7845.0    & 11.5      & 5640.6    & 11843.4     & 2606.4     & 5006.1     & 11211.4     & 12213.1     & 10746.1  & \textbf{13534.0} \\
            Seaquest        & 42054.7   & 68.4      & \textbf{683.3}  &    304.6     & 272.9     & 315.2     & 352.3     & 558.1     & 521.2   & 527.7 \\
            Up N Down       & 11693.2   & 533.4     & 3350.3    & 3075.0     & 2331.7     & 2646.4     & 4324.5     & 10859.2    & \textbf{14674.1}  & 10225.2 \\
            \bottomrule
        \end{tabular}
        \end{adjustbox}
    \caption{Scores achieved by different methods on Atari-100K. \textdagger~ denotes using virtual trajectories.}  
\label{table:atari_compare_entry}
\end{table*}

\end{document}